\documentclass[letterpaper]{article} 
\usepackage[preprint]{aaai2027}  
\usepackage[hyphens]{url}  
\usepackage{graphicx} 
\usepackage{natbib}  
\usepackage{caption} 
\usepackage{algorithm}
\usepackage{algorithmic}

\usepackage{newfloat}
\usepackage{listings}
\DeclareCaptionStyle{ruled}{labelfont=normalfont,labelsep=colon,strut=off} 
\floatstyle{ruled}
\newfloat{listing}{tb}{lst}{}
\floatname{listing}{Listing}

\usepackage{booktabs}

\usepackage{bm}
\usepackage{amssymb}
\usepackage{amsmath}
\usepackage{xcolor}

\newcommand{\revised}[1]{#1}
\title{\revised{Beyond Routing Saturation: A Long-Horizon Class-Incremental Perspective on Expert Routing in Multimodal Continual Instruction Tuning}}
\author{
    Huiyu Yi\textsuperscript{\rm 1,\rm 2},
    Yongqi Xu\textsuperscript{\rm 1,\rm 2}\equalcontrib,
    Bogang Zhang\textsuperscript{\rm 1,\rm 2}\equalcontrib,
    Dunwei Tu\textsuperscript{\rm 1,\rm 2},
    Zhiming Xu\textsuperscript{\rm 1,\rm 2},\\
    Zhen-Hao Xie\textsuperscript{\rm 1,\rm 2},
    Baile Xu\textsuperscript{\rm 1,\rm 2}\corresponding,
    Furao Shen\textsuperscript{\rm 1,\rm 2}\corresponding
}
\affiliations{
    \textsuperscript{\rm 1}School of Artificial Intelligence, Nanjing University, China\\
    \textsuperscript{\rm 2}National Key Laboratory for Novel Software Technology, Nanjing University, China\\
    huiyuyi@smail.nju.edu.cn,
    \{xubaile, frshen\}@nju.edu.cn
}

\begin{document}

\maketitle

\begin{abstract}
Multimodal Continual Instruction Tuning (MCIT) enables multimodal large
language models to acquire new tasks sequentially while retaining previously
learned capabilities. Many recent methods maintain task-specific LoRA experts
and route each input to one or more experts at inference. Yet the
task-identification problem underlying expert routing remains under-explored.
We show that routing is nearly saturated on widely used MCIT benchmarks.
Textual fingerprints that leak task identity and short 4--10-task sequences
with few competing experts jointly obscure the long-horizon routing problem.
To expose this challenge, we introduce FLEX (Fingerprint-reduced Long-horizon
Expert eXamination), a 34-task long-horizon MCIT benchmark with weakened
textual fingerprints. FLEX groups tasks with similar
instruction and answer formats but diverse visual and knowledge domains,
normalizes their outer templates, and evaluates routing over a substantially
larger expert pool. Crucially, we formulate progressive-LoRA routing as
soft task-as-class Multimodal Class-Incremental Learning (MCIL):
each task defines an incremental routing class, whose complete score
distribution supplies the LoRA mixture weights, with hard routing as a
discrete special case. FLEX exposes this expanding task-identification
challenge, while the MCIL formulation provides a principled interface for
transferring CIL methods to expert routing. We instantiate PureLoRA as a
controlled baseline and adapt four CIL methods to four MCIT frameworks without
modifying their LoRA experts or generation pipelines. Our plug-in routers
improve strict LoRA matching by up to 16.3 percentage points and overall
MacroScore by up to 4.6 points.

\end{abstract}


{
\begin{figure}[!h]
    \centering
    \includegraphics[width=\columnwidth]{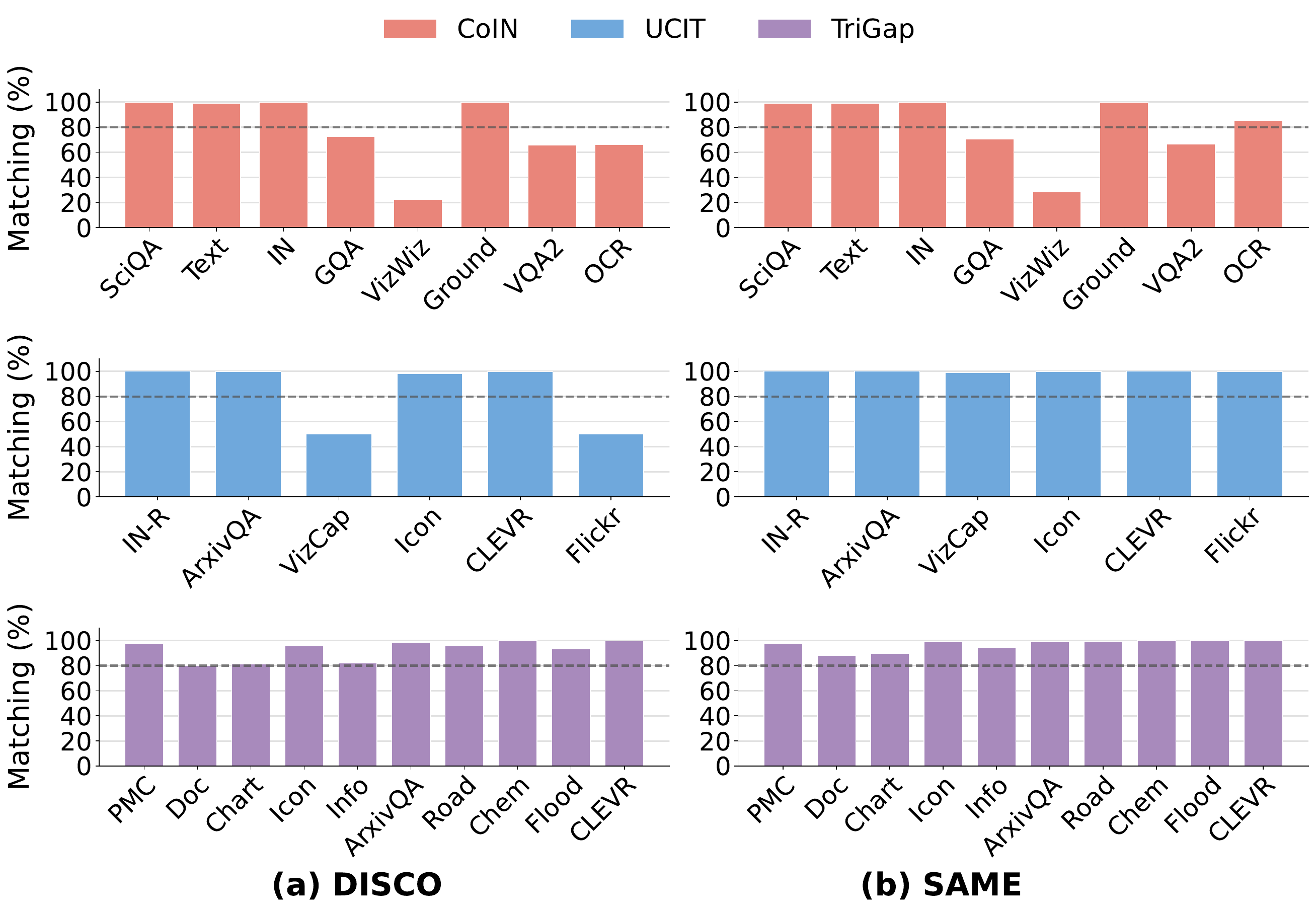}
    \caption{\revised{Task--LoRA matching on CoIN, UCIT, and TriGap. DISCO and SAME exceed 80\% matching on 17/24 and 21/24 tasks, respectively. Tied UCIT tasks under DISCO are shown by their 50/50 soft weights.}}
    \label{fig:textual_shortcuts}
\end{figure}
}

\section{Introduction}

Multimodal Continual Instruction Tuning (MCIT) aims to enable Multimodal
Large Language Models to acquire new tasks sequentially while retaining
previously learned capabilities~\citep{llava,zhang2026instruction-survey}.
Recent methods commonly maintain specialized LoRA experts and, at inference,
first estimate weights over historical LoRAs before generation. MoELoRA,
HiDe-LLaVA, DISCO, and SAME adopt or involve this
routing--composition--inference paradigm~\citep{coin-moelora,hidellava-ucit,disco,same-trigap}.
Yet the difficulty and scalability of routing remain under-explored.

We observe that routing is already close to saturation on widely used
benchmarks. Figure~\ref{fig:textual_shortcuts} shows that SAME reaches 99.76\%
overall task--LoRA matching on UCIT and 96.07\% on TriGap. DISCO exceeds 80\%
matching on 17 of 24 tasks across CoIN, UCIT, and TriGap despite routing
entirely from text; SAME is likewise text-dominant, fusing textual and visual
signals at an 8:2 ratio. When task-specific LoRAs isolate most updates and
task identity is nearly perfectly recovered, task-agnostic inference
approaches task-aware expert selection, leaving routing weakly tested.

We identify two benchmark properties that jointly contribute to this routing
saturation. First, existing tasks contain strong \emph{textual fingerprints}:
superficial instruction patterns, such as unique interfaces, fields, and
templates, that reveal dataset identity without image understanding, akin to
shortcut learning~\citep{shortcut}. Second,
CoIN, UCIT, and TriGap contain only 8, 6, and 10 tasks, exposing routers to
few competing LoRAs. Routing evaluation therefore requires both weak
identity leakage and a longer sequence of potentially confusable tasks.

To this end, we introduce \revised{Fingerprint-reduced Long-horizon Expert eXamination
(FLEX)}, a 34-task long-horizon MCIT benchmark with weakened textual
fingerprints. \revised{FLEX} groups datasets with
similar task interfaces and answer formats but diverse visual domains and
knowledge requirements, normalizes outer templates, and expands the historical
LoRA pool. Together, these properties make expert routing more challenging
and reveal an expanding task-identification problem.

More importantly, we observe that the routing stage can be naturally formulated as a
task-as-class Multimodal Class-Incremental Learning (MCIL)
problem~\citep{rebuffi2017icarl,ding2022don-clip}. Each newly arriving MCIT
task introduces a corresponding LoRA, and all samples from that task share
the same label in the routing space. Without task identity at test time, the
router must identify the appropriate LoRA among all previously learned tasks,
which directly resembles an expanding class-incremental classification
problem. The key distinction is that standard MCIL typically applies
$\arg\max$ to produce a single class prediction, whereas MCIT usually retains
the complete class-score distribution as soft weights over LoRA experts. We
therefore formulate MCIT routing as soft task-as-class MCIL. From this
perspective, the widely used combination of frozen CLIP image-text encoders,
task-wise mean prototypes, and similarity matching corresponds to a basic
frozen-feature incremental classifier, leaving substantial room for
improvement.

Based on this insight, we propose PureLoRA, a controlled
baseline that retains only task-specific LoRA training, multimodal task identification, and soft LoRA composition. We
further adapt HC, HC-SOINN, RanPAC, and DDAS~\citep{hc-hcsoinn,ranpac,DDAS} to the routers of
PureLoRA, HiDe-LLaVA, DISCO, and SAME without modifying their learned experts or generation pipelines. Experiments demonstrate consistent improvements in routing quality and downstream performance, with strict LoRA matching
gains of up to 16.3 percentage points and MacroScore gains of up to 4.6 points.

Our main contributions are summarized as follows:
\begin{itemize}
\item We reveal and diagnose expert-routing saturation in existing MCIT
benchmarks, and introduce \revised{FLEX} for more challenging multimodal LoRA-routing evaluation.
\item We reveal that MCIT routing
under the progressive-LoRA paradigm can be reformulated as soft task-as-class
MCIL, establishing a direct connection between expert routing and multimodal
class-incremental learning.
\item We propose the controlled PureLoRA baseline and adapt
multiple CIL methods to diverse MCIT routers, substantially improving both routing quality and downstream performance.
\end{itemize}

\section{Related Work}

\subsection{Multimodal Continual Instruction Tuning}

MCIT enables MLLMs to acquire vision--language capabilities
sequentially~\citep{llava,longpre2023flan,zhang2026instruction-survey}.
CoIN~(8), UCIT~(6), TriGap~(10), and MLLM-CL's ACL/DCL~(4 each) evaluate this
setting across heterogeneous tasks and domain
gaps~\citep{coin-moelora,hidellava-ucit,same-trigap,zhao2025mllm}. Their limited
horizons leave expert-routing scalability under a substantially expanding
task space under-examined. MoELoRA gates shared experts, SAME stabilizes
expert routing and updates, ProgLoRA freezes task-specific LoRAs, HiDe-LLaVA
separates shared and task-specific knowledge, and DISCO combines LoRAs through
input--prototype similarity~\citep{coin-moelora,same-trigap,proglora,hidellava-ucit,disco}.
Shared experts promote transfer but risk interference, whereas task-specific
experts depend on accurate routing as the candidate pool grows.

\subsection{Class-Incremental and Multimodal Class-Incremental Learning}

CIL learns new classes without test-time task identities~\citep{de2021continual,rebuffi2017icarl,mqmk,SAAN}, while MCIL additionally exploits multimodal representations~\citep{ding2022don-clip}. Their classifiers naturally support task-as-class routing: RanPAC uses a closed-form classifier over randomly projected features~\citep{ranpac}; DDAS routes by task-specific reconstruction errors~\citep{DDAS}; and HC-SOINN represents complex distributions with adaptive prototype graphs~\citep{hc-hcsoinn}. We adapt these mechanisms to LoRA routing.

\section{Preliminaries}

\subsection{Multimodal Continual Instruction Tuning.}
We consider a sequence of $T$ tasks. Task $t$ provides
$\mathcal{D}_t=\{(\mathbf{v}_i^t,\mathbf{q}_i^t,\mathbf{y}_i^t)\}_{i=1}^{N_t}$,
where $\mathbf{v}$, $\mathbf{q}$, and $\mathbf{y}$ denote the image,
instruction, and response. In the rehearsal-free setting, the model learns
only from $\mathcal{D}_t$ at stage $t$ using the standard autoregressive
objective, and must subsequently handle all seen tasks without task identities.

\subsection{Expert-based MCIT Paradigms.}
Let $W_\ell^0$ denote the pretrained weight of the $\ell$-th layer in the
MLLM. The $m$-th LoRA expert introduces a low-rank update
$\Delta W_{\ell,m}=B_{\ell,m}A_{\ell,m}$ at this layer. After learning task
$t$, let $M_t$ denote the number of available experts. Given an input
$\mathbf{x}$, the routing module produces
$\boldsymbol{\alpha}_t(\mathbf{x}) \in \mathbb{R}^{M_t}$, where
$\alpha_{t,m}(\mathbf{x})\geq 0$ and
$\sum_{m=1}^{M_t}\alpha_{t,m}(\mathbf{x})=1$. The resulting layer output can
be generally written as
\begin{equation}
\mathbf{h}_\ell =
W_\ell^0\mathbf{z}_\ell +
\sum_{m=1}^{M_t}
\alpha_{t,m}(\mathbf{x})
\Delta W_{\ell,m}\mathbf{z}_\ell,
\end{equation}
where $\mathbf{z}_\ell$ is the input to the $\ell$-th layer.

Existing methods mainly adopt two types of expert pools.
In a \emph{fixed expert pool}, the number of experts remains constant,
i.e., $M_t=M$, and different tasks share and continually update the same set
of experts~\citep{moe}. Representative methods include MoELoRA and SAME.
In a \emph{progressive LoRA pool}, each incoming task introduces a new
task-specific LoRA, such that $M_t=t$. Previously learned LoRAs are typically
preserved and are selected or fused at inference time according to the input.
Representative methods include ProgLoRA, HiDe-LLaVA, and DISCO.
Our subsequent task-as-class reformulation focuses on task-agnostic routing in progressive LoRA pools, where the one-to-one correspondence between tasks and LoRAs provides the foundation for formulating the routing stage as an MCIL problem.

\subsection{Evaluation Metrics.}
\label{sec:evaluation_metrics}
{
To make the distinction between downstream task quality and routing quality
explicit, we use the following metrics throughout the paper.
}
Let $A_{s,T}$ denote the final score on task $s$ after learning all $T$
tasks. Let $\boldsymbol{\alpha}_{s,i}^{(T)}\in\mathbb{R}^{T}$ be the
normalized LoRA-weight vector for the $i$-th evaluation sample of task $s$,
and let $N_s$ be the number of evaluation samples. We report the MacroScore
$\mathcal{M}$, Macro Ground-Truth Weight $\mathcal{G}$, and Overall Matching
Rate $\mathcal{R}$:
\begin{equation}
\begin{aligned}
\mathcal{M}
&=
\frac{1}{T}\sum_{s=1}^{T}A_{s,T},\\
\mathcal{G}
&=
\frac{1}{T}\sum_{s=1}^{T}g_s,
&\quad
g_s
&=
\frac{1}{N_s}\sum_{i=1}^{N_s}\alpha_{s,i,s}^{(T)},\\
\mathcal{R}
&=
\frac{\sum_{s=1}^{T}N_s r_s}{\sum_{s=1}^{T}N_s},
&\quad
r_s
&=
\frac{1}{N_s}\sum_{i=1}^{N_s}
\mathbb{I}\!\left[
\arg\max_k\alpha_{s,i,k}^{(T)}=s
\right].
\end{aligned}
\label{eq:evaluation_metrics}
\end{equation}
For hard-routing methods, $\boldsymbol{\alpha}_{s,i}^{(T)}$ is one-hot, and
the same definitions remain applicable.
We use high values of both $\mathcal{R}$ and $\mathcal{G}$ across a benchmark
as evidence of routing saturation, while their gaps to ground-truth routing
quantify the unresolved task-identification difficulty.


\section{\revised{FLEX: A Long-Horizon Benchmark with Weak Textual Fingerprints}}
\label{sec:flex}





\subsection{\revised{Textual Fingerprints and Routing Saturation}}

A \emph{textual fingerprint} is an instruction pattern that reveals dataset
identity without requiring image understanding or core question semantics.
We identify three common forms: task-interface cues, such as bounding-box or
classification outputs; dataset-specific templates or fields, such as
``\texttt{Reference OCR token:}''; and stable structural cues, including
question length and option layout. When a distinctive interface occurs in
only one dataset, text-only or text-dominant routers can infer its LoRA
directly from the prompt. The short 4--10-task horizons of existing benchmarks
further limit the number of competing LoRAs, but simply adding uniquely
tagged tasks would preserve the same shortcut. FLEX therefore primarily
weakens textual identity leakage and uses a longer sequence to evaluate the
resulting task ambiguity over a larger expert pool.

\subsection{\revised{Benchmark Construction and Composition}}
FLEX is designed to prevent shortcut-driven routing saturation rather than to
make matching arbitrarily low. It groups at least two potentially confusable
datasets under each task format, normalizes their outer templates while
retaining necessary content, and spans diverse visual and knowledge domains.
Its 34 tasks comprise 5 ImageNet-200 classification tasks, 2 captioning
datasets, 7 multiple-choice datasets, and 20 short-answer VQA datasets.
Shared interfaces weaken identity leakage, while the longer sequence expands
the historical LoRA pool. Table~\ref{tab:tfp_composition} summarizes FLEX.




\begin{table}[t]
    \centering
    \small
    \setlength{\tabcolsep}{4pt}
    \begin{tabular}{@{}lcl@{}}
        \toprule
        \textbf{Task Format} & \textbf{Tasks} &
        \textbf{Representative Datasets} \\
        \midrule
        Classification & 5 & ImageNet200\_1--5 \\
        Captioning & 2 & Vizcap, Flickr30k \\
        Multiple choice & 7 & ArxivQA, ScienceQA, PMCVQA, AI2D \\
        Short-answer VQA & \revised{20} & GQA, DocVQA, ChemVQA, PathVQA \\
        \bottomrule
    \end{tabular}
    \caption{\revised{A compact overview of the 34 continual tasks in FLEX.}}
    \label{tab:tfp_composition}
\end{table}

{\color{black}
The complete per-task catalog, including domain, answer style, retained
prefixes or special fields, frozen answer-format suffixes, and dataset
citations, is provided in Appendix~S3.7.
}

For sufficiently large datasets, we sample 10,000 training and 3,000 test
examples. Each ImageNet subset uses 30,000 training examples, while smaller
datasets retain all available samples. FLEX contains 388,856 training and
85,037 test examples and preserves each dataset's native metric.


\subsection{\revised{Does FLEX Avoid Routing Saturation?}}

\begin{figure}[!t]
    \centering
    \includegraphics[width=\columnwidth]{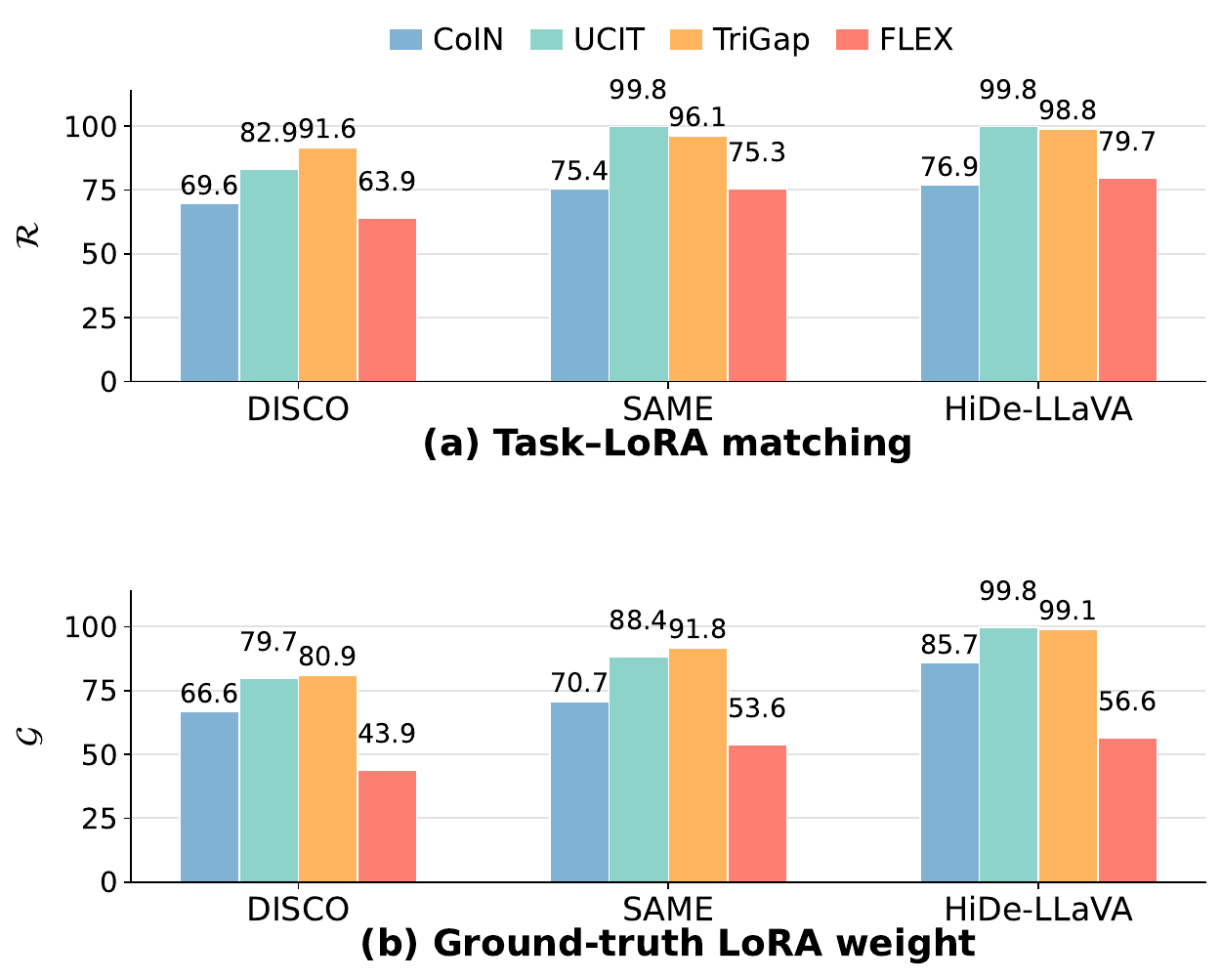}
    \caption{\revised{Routing quality across established MCIT benchmarks and FLEX. FLEX lowers ground-truth LoRA weight $\mathcal{G}$ in all nine comparisons and strict matching $\mathcal{R}$ in eight of nine.}}
    \label{fig:tfp_routing_gap}
\end{figure}

We compare DISCO, SAME, and HiDe-LLaVA across established benchmarks and FLEX
using strict task--LoRA matching $\mathcal{R}$ and macro ground-truth LoRA
weight $\mathcal{G}$. As shown in Figure~\ref{fig:tfp_routing_gap}, FLEX lowers
$\mathcal{G}$ in all nine measured established-benchmark--FLEX comparisons and
lowers $\mathcal{R}$ in eight of nine. The sole strict-matching exception is
HiDe-LLaVA on CoIN, whereas its ground-truth weight still decreases on FLEX.
The consistent reduction in $\mathcal{G}$ shows that the correct LoRA receives
less dominant soft-routing mass.
A controlled suffix-fingerprint injection further isolates the textual factor
and restores easy matching (Appendix~S3.8).
FLEX is not entirely fingerprint-free, but it weakens the dominant interface-to-identity shortcut while exposing a larger expert pool.

{
\begin{figure*}[!t]
    \centering
    \includegraphics[width=\textwidth]{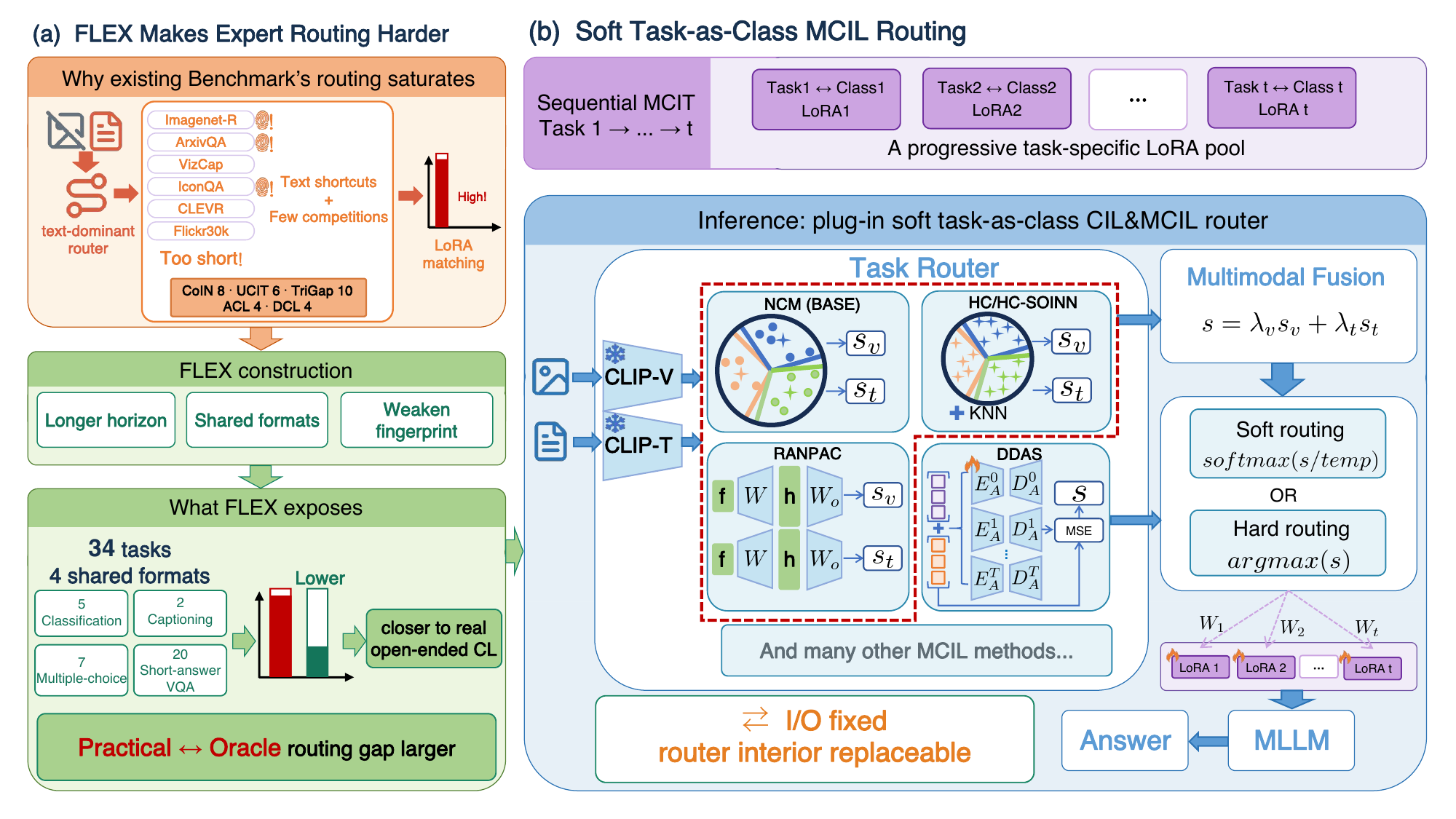}
    \caption{Overview of \revised{FLEX} and the proposed task-as-class MCIL formulation
    for multimodal LoRA routing.}
    \label{fig:tfp_mcil_overview}
\end{figure*}
}

\section{Reformulating MCIT Routing as MCIL}
\label{sec:reformulation}

\subsection{Soft Task-as-Class MCIL}

FLEX's long horizon makes the expanding structure of progressive-LoRA routing
explicit. In a progressive pool, task $k$ corresponds to LoRA $k$. We discard
the target response and assign every input
$\mathbf{x}_i^k=(\mathbf{v}_i^k,\mathbf{q}_i^k)$ the routing label
$c_i^k=k$, yielding

\begin{equation}
\widetilde{\mathcal{D}}_k
=
\{(\mathbf{x}_i^k,c_i^k)\}_{i=1}^{N_k},
\qquad c_i^k=k.
\end{equation}
After task $t$, the routing label space is
$\mathcal{C}_t=\{1,\ldots,t\}$. Each new task thus enlarges both the class
space and the candidate LoRA pool. Since task identity is unavailable at
inference, routing becomes an expanding task-classification problem, matching
the setting of Multimodal Class-Incremental Learning (MCIL). This
correspondence is especially consequential under long horizons, where a
router must remain comparable and calibrated across an increasing number of
historical tasks.



Let $G_t$ denote the router at stage $t$, which outputs
\begin{equation}
G_t(\mathbf{x})
=
\mathbf{s}_t(\mathbf{x})
=
\left[
s_{t,1}(\mathbf{x}),
\ldots,
s_{t,t}(\mathbf{x})
\right]
\in\mathbb{R}^{t}.
\end{equation}
A hard-routing method predicts
\begin{equation}
\widehat{c}_t(\mathbf{x})
=
\arg\max_{k\in\mathcal{C}_t}
s_{t,k}(\mathbf{x})
\end{equation}
and activates only the corresponding LoRA. In this case, MCIT routing has the
same classification decision form as standard MCIL. Its routing weights can
be written as
\begin{equation}
\alpha_{t,k}^{\mathrm{hard}}(\mathbf{x})
=
\mathbb{I}
\left[
k=\widehat{c}_t(\mathbf{x})
\right].
\end{equation}

A soft-routing method instead normalizes the task scores using
\begin{equation}
\alpha_{t,k}^{\mathrm{soft}}(\mathbf{x})
=
\frac{
\exp\left(\tau\,s_{t,k}(\mathbf{x})\right)
}{
\sum_{j=1}^{t}
\exp\left(\tau\,s_{t,j}(\mathbf{x})\right)
},
\label{eq:soft_routing}
\end{equation}
where $\tau>0$ is a logit scale~\citep{temperature}
(i.e., PureLoRA uses $\tau{=}28$). The resulting vector
\begin{equation}
\boldsymbol{\alpha}_t(\mathbf{x})
=
\left[
\alpha_{t,1}(\mathbf{x}),
\ldots,
\alpha_{t,t}(\mathbf{x})
\right]
\end{equation}
is used in the LoRA composition defined in the preliminaries. Therefore,
hard routing is directly equivalent to ordinary incremental class prediction,
whereas soft routing is a continuous relaxation that retains confidence over
all task classes for LoRA composition. We refer to this unified view as
\textbf{soft task-as-class MCIL}, with hard routing as its discrete special
case.

This direct correspondence mainly applies to progressive LoRA pools. In a
fixed expert pool, such as MoELoRA or SAME, task identities and expert indices
are not one-to-one, and an additional task-to-expert mapping is required.

\subsection{Existing Routing as Frozen-Feature NCM}

Existing progressive-LoRA routers commonly use frozen CLIP encoders $f_v$ and $f_q$. For task $k$, they store visual and textual mean prototypes
\begin{equation}
\boldsymbol{\mu}_k^m
=
\frac{1}{N_k}\sum_{i=1}^{N_k} f_m(\mathbf{x}_i^{m,k}),
\qquad m\in\{v,q\}.
\end{equation}
For a test input, cosine similarities to these prototypes are fused as
\begin{equation}
s_{t,k}(\mathbf{x})
=
\lambda\,\operatorname{sim}(f_v(\mathbf{v}),\boldsymbol{\mu}_k^v)
+
(1-\lambda)\,\operatorname{sim}(f_q(\mathbf{q}),\boldsymbol{\mu}_k^q).
\end{equation}


where $\lambda\in[0,1]$ controls the contribution of the visual modality.
The resulting $\mathbf{s}_t(\mathbf{x})$ is used for either hard LoRA
selection or soft LoRA weighting.

From an MCIL perspective, this procedure corresponds to a nearest-class-mean
(NCM) classifier~\citep{ncm} over frozen multimodal features: each continual task is
treated as an incremental class and represented by its mean prototypes.
Although methods such as HiDe-LLaVA and DISCO differ in modality weights,
applied layers, and LoRA composition strategies, their first-stage routing can
be summarized by this basic incremental classification paradigm.

Frozen CLIP encoders with NCM routing are only one simple instantiation of
MCIL-based routing. More generally, any multimodal incremental learning method
that produces comparable scores over an expanding set of tasks may serve as
an MCIT router.

\subsection{CIL-based Routing Enhancement}

Under the task-as-class MCIL formulation, we instantiate four incremental
task-identification methods, namely HC, HC-SOINN,
RanPAC, and DDAS, and integrate them into the first-stage
routers of different MCIT frameworks. All methods identify the historical
task associated with the current input and accordingly select or weight the
LoRA experts, while leaving the learned LoRAs and the subsequent generation
pipeline unchanged. To control additional computation and isolate the effect
of routing, our current implementations freeze both CLIP encoders and use
their outputs only as routing features.

For HC, HC-SOINN, and RanPAC, we maintain a task identifier for each
modality:
\begin{equation}
\mathbf{s}_t^v
=
G_t^v\!\left(f_v(\mathbf{v})\right),
\qquad
\mathbf{s}_t^q
=
G_t^q\!\left(f_q(\mathbf{q})\right),
\end{equation}
where $\mathbf{s}_t^v,\mathbf{s}_t^q\in\mathbb{R}^{t}$ denote the task-score
vectors produced by the visual and textual branches. Their scores are fused
using the modality ratio of the target MCIT framework:
\begin{equation}
\mathbf{s}_t
=
\lambda\mathbf{s}_t^v
+
(1-\lambda)\mathbf{s}_t^q.
\label{eq:cil_fusion}
\end{equation}

\textbf{HC} constructs multiple hierarchical cluster
prototypes~\citep{hc,hc-hcsoinn} for each
task and computes routing scores according to cosine similarities between an
input and the task prototypes. \textbf{HC-SOINN} dynamically grows and
updates prototype nodes through an online self-organizing network, and
identifies tasks based on node distances and local densities.
\textbf{RanPAC} applies a fixed random projection to the routing features and
produces task logits using a closed-form ridge-regression classifier. HC and
HC-SOINN do not train an explicit router through backpropagation, while
RanPAC only updates and solves a closed-form classifier.

\textbf{DDAS} adopts a different multimodal design. It first concatenates the
outputs of the frozen CLIP dual encoders:
\begin{equation}
\mathbf{z}^{\mathrm{cat}}
=
\left[
f_v(\mathbf{v});
f_q(\mathbf{q})
\right].
\end{equation}
An autoencoder $A_k$ is maintained for each historical task, and the routing
score for task $k$ is defined as the negative reconstruction error:
\begin{equation}
s_{t,k}^{\mathrm{DDAS}}(\mathbf{x})
=
-\kappa\,
\operatorname{MSE}
\left(
\mathbf{z}^{\mathrm{cat}},
A_k(\mathbf{z}^{\mathrm{cat}})
\right),
\label{eq:ddas_score}
\end{equation}
where $\kappa$ is a score-scaling factor. DDAS therefore performs task
identification using a single autoencoder bank over concatenated multimodal
features, rather than separate visual and textual routers followed by linear
fusion. Unlike the other three methods, DDAS trains one lightweight
autoencoder for each incoming task.
The resulting task scores are converted into LoRA routing decisions according
to the policy of the target MCIT framework.

Not all MCIL methods transfer directly because each entire MCIT task forms one
routing class. Methods requiring intra-task semantic classes or inter-class
relations need redefined supervision, whereas distribution models that produce
comparable scores over historical tasks are directly compatible. Implementation
details and broader applicability are discussed in Appendix~S2.5.

{
Figure~\ref{fig:tfp_mcil_overview} summarizes the proposed benchmark design
and the task-as-class MCIL view of multimodal LoRA routing.
}

\section{Experiments}
\subsection{Experimental Setup}
\label{sec:experimental_setup}

\paragraph{Benchmark and Methods.}
We evaluate all historical tasks after completing the 34-task FLEX sequence.
We compare HiDe-LLaVA, DISCO, and SAME, and additionally include
\textbf{PureLoRA}, a minimal progressive-LoRA baseline containing only
independent task experts, multimodal task identification, and soft expert
composition. Its full design and implementation are provided in Appendix S1.
HC, HC-SOINN, RanPAC, and DDAS are separately integrated into the first-stage
router of each framework, resulting in 16 routing-enhanced variants while
leaving the learned experts and generation pipelines unchanged. PureLoRA,
DISCO, SAME, and HiDe-LLaVA all use soft routing; HiDe-LLaVA applies soft
\texttt{delta\_mean} remain composition.

\paragraph{Implementation Details.}
All methods are implemented under PRISM~\cite{prism} and follow SAME's training protocol
unless otherwise specified. We use LLaVA-v1.5-7B~\cite{llava} as the backbone
and CLIP-L/14-336~\cite{clip} for visual and textual routing features. SAME
inserts LoRA modules into all language-model linear layers, whereas PureLoRA,
HiDe-LLaVA, and DISCO use the attention and FFN linear layers; all task
experts have rank 8. Each task is trained for one epoch with AdamW and a
learning rate of $2\times10^{-4}$. SAME uses total MoE rank 272 in the
34-task setting. Experiments run on 8 T-Head Zhenwu 810E PPUs.

\begin{table*}[t]
    \centering
    \small
    \setlength{\tabcolsep}{4pt}
    \begin{tabular}{llccccccc}
    \toprule
    MCIT & Router & Classification & Captioning & MCQ & VQA & $\mathcal{M}$ & $\mathcal{G}$ & $\mathcal{R}$ \\
    \midrule
    DISCO  & Base & 44.85 & 51.57 & 70.30 & 44.56 & 50.32 & 43.90 & 63.95 \\
    (fixed)& HC & 44.61 & 51.52 & \textbf{72.40} & 46.33 & 51.75\,(+1.43) & 51.18 & 64.58 \\
     & HC-SOINN & \underline{45.03} & \underline{51.58} & 71.22 & \underline{46.81} & 51.86\,(+1.54) & 59.63 & 66.87 \\
     & DDAS & \textbf{59.51} & \textbf{59.02} & 71.64 & \textbf{47.52} & \textbf{54.93}\,(+4.61) & \textbf{78.86} & \textbf{80.21} \\
     & RanPAC & 44.94 & 51.55 & \underline{71.79} & \textbf{47.52} & \underline{52.37}\,(+2.06) & \underline{63.90} & \underline{70.37} \\
    \midrule
    SAME & Base & 47.99 & 51.70 & 67.79 & 44.38 & 50.16 & 53.65 & 75.30 \\
     & HC & 52.10 & 51.79 & \underline{70.14} & \underline{46.34} & 52.41\,(+2.24) & 65.20 & 81.03 \\
     & HC-SOINN & \underline{56.83} & 51.81 & 69.79 & 45.88 & 52.76\,(+2.60) & 66.19 & \underline{84.97} \\
     & DDAS & 54.75 & \textbf{57.03} & 69.95 & 46.05 & \underline{52.90}\,(+2.73) & \textbf{79.79} & 81.33 \\
     & RanPAC & \textbf{60.43} & \underline{56.81} & \textbf{70.72} & \textbf{46.62} & \textbf{54.21}\,(+4.05) & \underline{74.76} & \textbf{86.40} \\
    \midrule
    HiDe-LLaVA  & Base & 59.49 & \underline{45.81} & 58.49 & 35.54 & 44.39 & 56.59 & 79.69 \\
    (fixed)& HC & 59.81 & 45.80 & 58.48 & \textbf{35.64} & 44.50\,(+0.11) & 67.58 & 82.44 \\
     & HC-SOINN & 59.63 & 45.80 & \underline{58.54} & \underline{35.63} & 44.48\,(+0.09) & 64.60 & \underline{82.56} \\
     & DDAS & \underline{60.09} & \textbf{45.84} & \textbf{58.56} & 35.58 & \underline{44.52}\,(+0.13) & \textbf{79.46} & 80.21 \\
     & RanPAC & \textbf{60.83} & 45.80 & 58.52 & \textbf{35.64} & \textbf{44.65}\,(+0.26) & \underline{76.63} & \textbf{86.27} \\
    \midrule
    PureLoRA & Base & 57.04 & 57.02 & 72.42 & 47.51 & 54.60 & 60.81 & 83.77 \\
     & HC & 57.34 & 56.93 & \underline{72.92} & \textbf{48.16} & 55.12\,(+0.53) & 65.13 & 81.44 \\
     & HC-SOINN & 58.19 & 57.83 & 72.67 & \underline{48.14} & \underline{55.24}\,(+0.64) & 69.60 & \underline{83.92} \\
     & DDAS & \underline{59.51} & \textbf{59.03} & 72.53 & 47.62 & 55.17\,(+0.57) & \textbf{79.79} & 81.33 \\
     & RanPAC & \textbf{64.20} & \underline{59.00} & \textbf{73.34} & 47.82 & \textbf{56.14}\,(+1.54) & \underline{78.32} & \textbf{87.72} \\
    \bottomrule
    \end{tabular}
    \caption{Task-type results on the 34-task \revised{FLEX}. Category scores are unweighted task means;
parentheses denote changes from Base. Best/second-best within each MCIT block are bold/underlined. All 20 configurations are complete.
``(fixed)'' marks corrected HiDe-LLaVA and DISCO settings; their original ports performed abnormally poorly on FLEX.}
    \label{tab:main_results_task_types}
    \end{table*}
\subsection{Main Results}

\begin{figure*}[t]
    \centering
    \includegraphics[width=\textwidth]{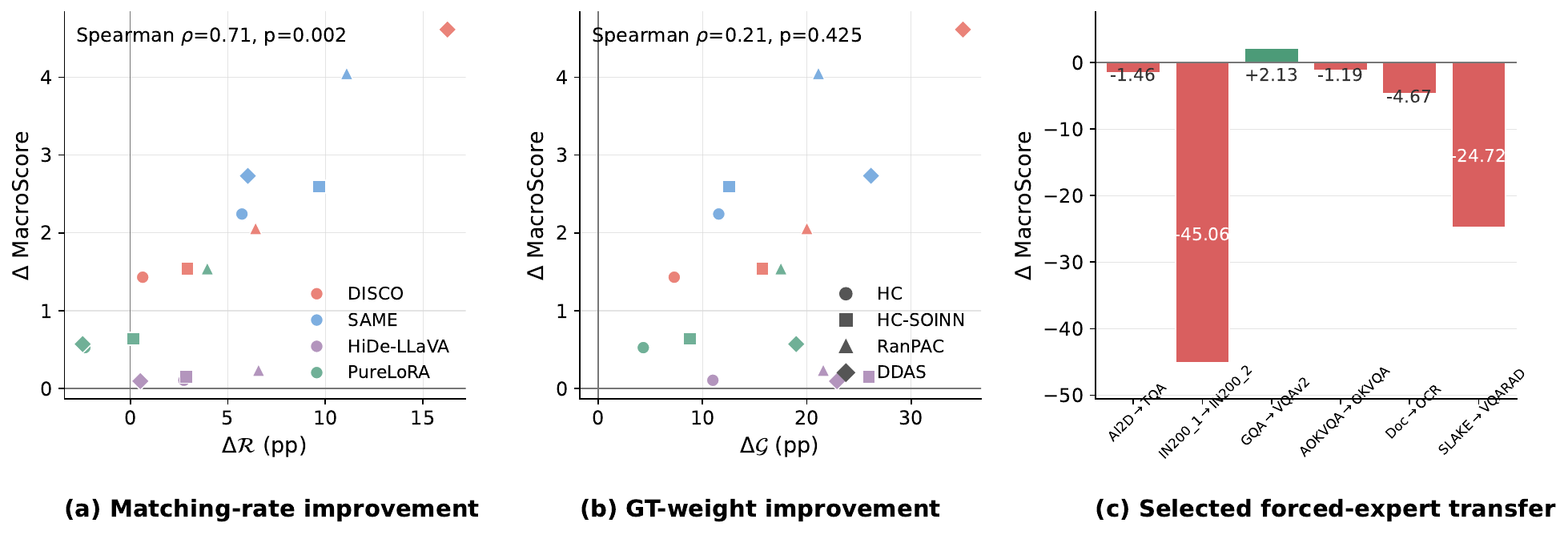}
    \caption{Routing quality and downstream performance. (a,b) Each point is one CIL router minus its Base: $\Delta\mathcal{M}$ versus $\Delta\mathcal{R}$ (top-1 accuracy) or $\Delta\mathcal{G}$ (correct-LoRA weight); $n=16$. (c) MacroScore change when six frequent PureLoRA confusions are forced in isolation.}
    \label{fig:e4_routing_performance}
\end{figure*}

{

Table~\ref{tab:main_results_task_types} reports task-type results on FLEX.
Unlike shorter MCIT settings where routing saturates, FLEX leaves a clear gap
between practical and ground-truth routing over the full expert pool.
Our plug-in routers alter only first-stage task identification and LoRA mixture
weights, while keeping experts, backbone, and generation fixed, so downstream
changes can be attributed to routing.

On the routing-sensitive frameworks DISCO, SAME, and PureLoRA, all 12
CIL transfers improve both $\mathcal{M}$ and $\mathcal{G}$ (12/12), and
$\mathcal{R}$ in 10/12 cases.
Mean MacroScore gains are 2.41 (DISCO), 2.91 (SAME), and 0.82 (PureLoRA).
The strongest routers raise DISCO from 50.32 to 54.93 with DDAS (+4.61),
SAME from 50.16 to 54.21 with RanPAC (+4.05), and PureLoRA from 54.60 to
56.14 with RanPAC (+1.54).
Prototype-based HC and HC-SOINN yield stable gains, while DDAS and RanPAC lead
different frameworks; 33 router--task improvements exceed 9 MacroScore points,
with a maximum of 30.96 for DISCO+DDAS on ImageNet200$_1$.
HiDe-LLaVA remains largely routing-insensitive in downstream score: its five
completed configurations span only 0.26 MacroScore despite much larger
routing-metric shifts.
Appendix~S2.1--S2.4 explain why PureLoRA attains the strongest Base
MacroScore, why SAME as a fixed expert pool still admits soft task-as-class
routing, and what the ``(fixed)'' DISCO and HiDe-LLaVA entries in
Table~\ref{tab:main_results_task_types} denote.
}

\begin{figure}[t]
    \centering
    \includegraphics[width=\columnwidth]{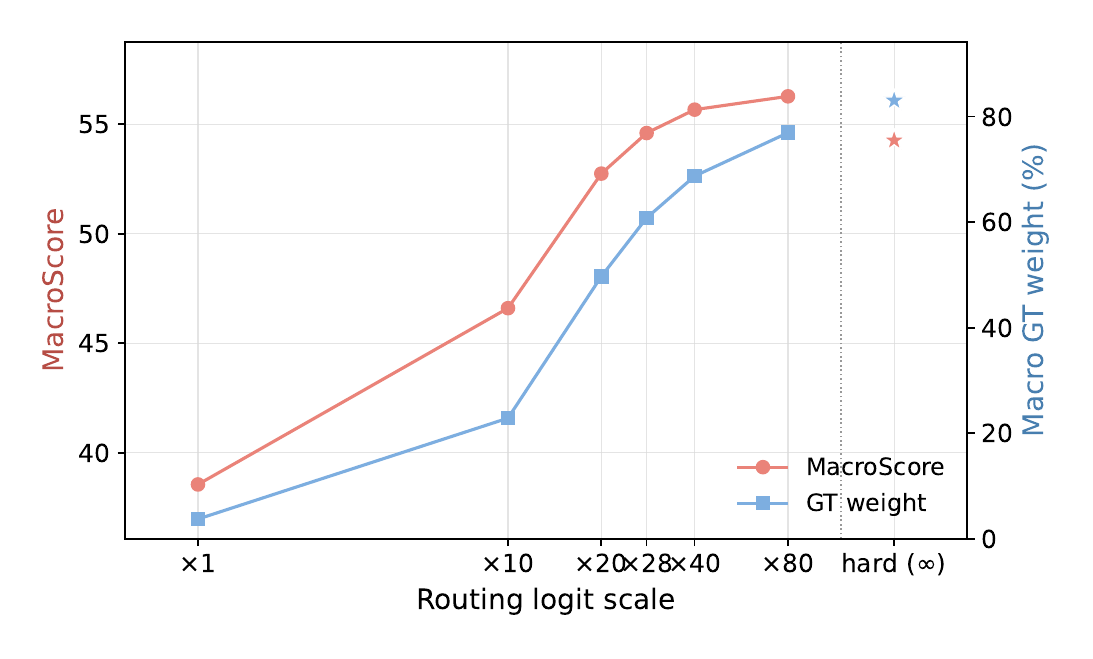}
    \caption{PureLoRA routing-scale calibration with fixed experts and prototypes. The $\times80$ soft router reaches the highest MacroScore and outperforms both the $\times28$ Base and hard routing.}
    \label{fig:e5_scale}
\end{figure}

\subsection{Further Analysis}

{
\paragraph{Closing the Oracle gap.}
\begin{table}[t]
    \centering
    \small
    \setlength{\tabcolsep}{3pt}
    \begin{tabular}{cccccc}
\toprule
Method & Base & Best CIL & Oracle & Gap & Recovered (\%) \\
\midrule
DISCO & 50.32 & 54.93\,(+4.61) & 60.01 & 5.08 & 47.6 \\
SAME & 50.16 & 54.21\,(+4.05) & 58.22 & 4.01 & 50.2 \\
HiDe-LLaVA & 44.39 & 44.65\,(+0.26) & 45.28 & 0.63 & 29.2 \\
PureLoRA & 54.60 & 56.14\,(+1.54) & 60.04 & 3.90 & 28.3 \\
\bottomrule
\end{tabular}

    \caption{FLEX Oracle-gap diagnostic. Parentheses show gain over Base.}
    \label{tab:oracle_gap}
\end{table}
We denote ground-truth one-hot routing as Oracle.
Table~\ref{tab:oracle_gap} reports the Base-to-Oracle MacroScore gap and the
fraction recovered by transferred CIL routers.
DISCO improves from 50.32 with its Base router to
54.93 with DDAS, compared with an Oracle score of 60.01, recovering 47.6\% of
the Base-to-Oracle gap. SAME improves from 50.16 to 54.21 with RanPAC against
an Oracle score of 58.22, recovering 50.2\%, while PureLoRA improves from
54.60 to 56.14 against 60.04, recovering 28.3\%. Therefore, the transferred
CIL routers remove a substantial fraction of task-identification error, but
the remaining gaps leave meaningful room for stronger routers. Oracle is a
routing reference rather than a guaranteed global upper bound, because a
compatible cross-task expert can occasionally provide positive transfer.

\paragraph{Routing quality and expert compatibility.}
Figure~\ref{fig:e4_routing_performance}(a,b) compares each of the 16 completed
CIL configurations with its framework-specific Base router. Each point plots
the MacroScore change $\Delta\mathcal{M}$ against a routing change:
$\Delta\mathcal{R}$ in~(a), i.e., how much more often the top-1 expert is
correct, and $\Delta\mathcal{G}$ in~(b), i.e., how much more probability mass
falls on the correct LoRA. Spearman $\rho$ measures whether larger routing
gains tend to accompany larger score gains across configurations ($\rho=1$:
perfect rank alignment; $\rho=0$: none); $p$ estimates how likely such
co-movement is under chance ($p<0.05$: unlikely to be random).

In aggregate, $\Delta\mathcal{M}$ co-moves with $\Delta\mathcal{R}$
($\rho=0.709$, $p=0.002$): routers that identify the correct expert more
often tend to score higher overall. The link to $\Delta\mathcal{G}$ is much
weaker ($\rho=0.215$, $p=0.425$), largely because HiDe-LLaVA can raise
correct LoRA weight without improving MacroScore. This pattern is not
task-wise guaranteed: task-level $\rho$ spans $0.188$--$0.827$ for
$\Delta\mathcal{R}$ and $0.164$--$0.830$ for $\Delta\mathcal{G}$.

Panel~(c) forces six frequent PureLoRA confusions in isolation. The wrong
expert underperforms the learned soft mixture in 5/6 cases (mean $-12.50$;
worst $-45.06$ for ImageNet200$_1\!\rightarrow$ImageNet200$_2$), while
GQA$\rightarrow$VQAv2 improves by $+2.13$.
Overall, routing quality predicts downstream gain mainly through better
top-1 identification; the harm of a specific mismatch depends on cross-task
LoRA compatibility rather than on routing metrics alone.

\paragraph{Soft-routing calibration.}
To isolate routing sharpness, we hold the PureLoRA checkpoint and router
prototypes fixed and vary only the positive logit scale before
softmax~\citep{temperature}. The
best soft endpoint, $\times80$, reaches $\mathcal{M}=56.27$ and
$\mathcal{G}=76.94$, exceeding the $\times28$ Base result ($\mathcal{M}=54.60$)
by 1.67 points. All soft scales retain the same $\mathcal{R}=83.77\%$, as
positive scaling does not change the top-ranked expert. The gain must therefore
arise from reweighting rather than reranking.

Hard top-1 routing yields the most concentrated allocation
($\mathcal{G}=83.04\%$, $\mathcal{R}=83.77\%$), yet obtains only
$\mathcal{M}=54.26$, 2.01 points below $\times80$. Selecting the predicted
expert exclusively is therefore not optimal even when top-1 identity is
unchanged. A calibrated soft mixture can preserve task specificity
while reusing compatible LoRAs from related tasks, giving it a higher
performance ceiling than hard routing.
}

\section{Conclusion}
{
We identify routing saturation as a hidden limitation of existing MCIT
evaluation: on 4--10-task benchmarks, textual fingerprints allow text-only or
text-dominant routers to recover task identity with near-perfect accuracy.
With task-specific LoRAs isolating most updates, task-agnostic inference
approaches task-aware expert selection, tracing the problem to two interacting
properties: textual identity leakage and short sequences with few competing
LoRAs.

To address both properties, we introduce FLEX, a 34-task long-horizon
benchmark with weakened textual fingerprints that removes shortcut-driven
saturation and tests routing over an expanding expert pool. We further reinterpret progressive-LoRA routing as soft
task-as-class MCIL, allowing established CIL classifiers to serve as plug-in
routers without retraining experts or changing generation.

Across three routing-sensitive frameworks, every transferred CIL method
improves MacroScore and ground-truth LoRA weight, and the best routers recover
28.3--50.2\% of the Oracle gap. Further analyses show that textual cues can
restore easy routing, mismatch costs depend on expert compatibility, and
calibrated soft mixtures can outperform hard selection; future MCIT systems
should therefore consider both long-horizon task identification and
compatibility-aware expert composition.

}


\bibliography{aaai2027}


\end{document}


\raggedbottom

\maketitle


\section{\revised{PureLoRA: A Controlled Progressive-LoRA Baseline}}
\label{sec:supp_purelora}

{\color{black}
\subsection{Design Goal}

PureLoRA is designed to isolate the effect of task identification and expert
composition in progressive-LoRA MCIT. It deliberately contains only three
components: independently trained task-specific LoRA experts~\citep{lora,proglora}, a frozen
multimodal prototype router, and soft LoRA composition. It does not introduce
a shared expert, a learned gating network, replay, or an auxiliary
continual-learning loss. Consequently, replacing its router changes neither
the expert pool nor the generation pipeline, making PureLoRA a controlled
testbed for comparing routing algorithms.

\subsection{Independent Task-Expert Training}

Let $W_\ell^0$ be a frozen linear transformation in layer $\ell$. After
learning $T$ tasks, PureLoRA stores one low-rank update
$\Delta W_{\ell,k}=B_{\ell,k}A_{\ell,k}$ for every task $k$. While learning
task $k$, only $(A_{\ell,k},B_{\ell,k})$ is trainable; the backbone and all
previous task experts remain frozen. The training-time forward pass therefore
uses a one-hot expert assignment. At inference, the layer output is
\begin{equation}
  \mathbf{h}_{\ell}
  =
  W_\ell^0\mathbf{z}_{\ell}
  +
  \sum_{k=1}^{T}
  \alpha_k(\mathbf{x})
  B_{\ell,k}A_{\ell,k}\mathbf{z}_{\ell},
  \qquad
  \sum_{k=1}^{T}\alpha_k(\mathbf{x})=1 .
\end{equation}
This separation prevents the parameters of an earlier task expert from being
overwritten by later tasks. Any remaining performance gap between inferred
and ground-truth routing can therefore be attributed primarily to expert
identification and composition rather than direct forgetting inside the
stored LoRAs.

\subsection{Multimodal Prototype Routing}

PureLoRA uses frozen CLIP image and text encoders~\citep{clip}, denoted by $f_v$ and
$f_q$. During task $k$, it updates one running mean prototype per available
modality:
\begin{equation}
  \boldsymbol{\mu}_k^v
  =
  \frac{1}{N_k^v}\sum_{i=1}^{N_k^v} f_v(\mathbf{v}_i^k),
  \qquad
  \boldsymbol{\mu}_k^q
  =
  \frac{1}{N_k^q}\sum_{i=1}^{N_k^q} f_q(\mathbf{q}_i^k).
\end{equation}
For a test input $\mathbf{x}=(\mathbf{v},\mathbf{q})$, its score for task $k$
is the equally weighted cosine similarity
\begin{equation}
  s_k(\mathbf{x})
  =
  \tfrac{1}{2}
  \operatorname{sim}
  \left(f_v(\mathbf{v}),\boldsymbol{\mu}_k^v\right)
  +
  \tfrac{1}{2}
  \operatorname{sim}
  \left(f_q(\mathbf{q}),\boldsymbol{\mu}_k^q\right).
  \label{eq:supp_purelora_score}
\end{equation}
If one modality is unavailable, the active modality weights are
renormalized. The base router converts the task scores into expert weights by
\begin{equation}
  \alpha_k(\mathbf{x})
  =
  \frac{\exp\!\left(\tau s_k(\mathbf{x})\right)}
  {\sum_{j=1}^{T}\exp\!\left(\tau s_j(\mathbf{x})\right)},
  \qquad \tau=28 .
  \label{eq:supp_purelora_softmax}
\end{equation}
The same $\boldsymbol{\alpha}(\mathbf{x})$ is applied to all adapted language
model layers. The default evaluation uses this soft mixture. Each task adds two mean routing prototypes and one LoRA expert, so
both routing memory and expert storage grow linearly with the number of
tasks.

\subsection{Router-Replacement Protocol}

Under the soft task-as-class MCIL view, the prototype similarities in
Equation~\ref{eq:supp_purelora_score} can be replaced by the task scores of
HC, HC-SOINN, RanPAC, or DDAS~\citep{hc-hcsoinn,ranpac,DDAS}. The resulting scores are normalized and used
as the coefficients in the same LoRA composition equation. Across all
PureLoRA variants, we keep the backbone, trained LoRA experts, decoding
procedure, and evaluation data fixed. Thus, Base-versus-CIL comparisons
measure the effect of the router rather than a change in expert capacity or
language generation.

\subsection{Implementation Details}

We use LLaVA-v1.5-7B as the backbone and CLIP ViT-L/14-336~\citep{llava,clip} for both routing
modalities. PureLoRA is inserted into the attention and FFN linear layers of
the language model only; the multimodal projector and backbone remain frozen.
Every task expert has rank 8. In the 34-task FLEX setting, these experts are
stored with aggregate $r=272$, $\mathtt{lora\_alpha}=544$, and dropout 0.05.
Each task is trained for one epoch with AdamW~\citep{adam}, a learning rate of
$2\times10^{-4}$, a warm-up ratio of 0.03, and cosine learning-rate decay.
These settings are shared by the Base and all routing-enhanced PureLoRA
variants.
}

\section{Additional Analyses for the Main Experiments}
\label{sec:supp_main_analyses}

\subsection{Why HiDe-LLaVA Is Routing-Insensitive, and What ``(fixed)'' Means}
\label{sec:supp_hide_insensitive}

The main experiment table shows that HiDe-LLaVA~\citep{hidellava-ucit} changes little in MacroScore when CIL routers improve task identification: $\mathcal{G}$ rises from $56.59$ to as high as $79.46$ and $\mathcal{R}$ from $79.69\%$ to as high as $86.27\%$, yet $\mathcal{M}$ stays within a $0.26$-point range ($44.39$--$44.65$).
This section explains (i) why the main-table HiDe row is marked ``(fixed)'', and (ii) why improved routing barely changes downstream performance once that protocol is in place.
We write $\mathcal{M}$, $\mathcal{G}$, and $\mathcal{R}$ as in Evaluation Metrics in the main text.

\paragraph{Architectural bottleneck.}
HiDe hierarchically decouples the language model: only the \emph{top} layer uses task-specific LoRA with prototype-based soft routing; all \emph{remain} layers share a single history-fused LoRA built from the experts seen so far~\citep{hidellava-ucit}.
In the protocol reported in our main table, remain layers use equal-weight fusion (\texttt{delta\_mean}, defined below) while the top layer uses soft routing $\mathrm{softmax}(\tau s)$ with $\tau{=}28$.
Under this design, remain always applies the same $1/N$ mixture over historical experts regardless of how sharply the top router concentrates on one task.
A better top match therefore has limited leverage over generation: even when the top layer often selects the correct expert---as suggested by HiDe's already-high Base $\mathcal{R}$ on FLEX ($79.69\%$) alongside the lowest Base $\mathcal{M}$ among the four frameworks---the remain path still averages every past expert into the forward pass.

\paragraph{What ``(fixed)'' means.}
The ``(fixed)'' tag denotes that our main-table HiDe numbers use a \emph{corrected remain-fusion and routing protocol}, not Prism's initial HiDe port nor an unnormalized reading of the official remain product.
Two implementation families diverge already at the remain layer.
Let $A_i\!\in\!\mathbb{R}^{r\times d}$ and $B_i\!\in\!\mathbb{R}^{d\times r}$ denote the LoRA factors of task expert $i$ (width $d$, rank $r$), and let $N$ be the number of observed tasks at evaluation ($N{=}34$ after the full FLEX sequence).
The \emph{official HiDe source code} fuses factors before a single application (\texttt{ab\_sum}):
\begin{equation}
  \Delta_{\mathrm{ab\_sum}}(x)
  =
  \Bigl(\sum_{i=1}^{N} B_i\Bigr)
  \Bigl(\sum_{i=1}^{N} A_i\Bigr) x ,
  \label{eq:hide_ab_sum}
\end{equation}
which expands to all cross terms $B_iA_j$.
\emph{Prism's initial port} instead summed per-expert updates (\texttt{delta\_sum}):
\begin{equation}
  \Delta_{\mathrm{delta\_sum}}(x)
  =
  \sum_{i=1}^{N} B_i(A_i x) ,
  \label{eq:hide_delta_sum}
\end{equation}
i.e., only the self terms, with no explicit cross-expert products.
The two forms are algebraically inequivalent: Eq.~\eqref{eq:hide_ab_sum} fuses weights then applies once, whereas Eq.~\eqref{eq:hide_delta_sum} applies each expert and sums the outputs (mathematically $\sum_i B_iA_ix$, implemented as $\sum_i B_i(A_ix)$).
That initial port also used hard top-1 routing; under \texttt{delta\_sum}+hard, FLEX Base collapses to $\mathcal{M}{=}7.96$ (Table~\ref{tab:hide_remain_fuse}, row~1).

\paragraph{Which form matches the HiDe paper?}
HiDe-LLaVA describes remain-layer fusion at the \emph{module} level: Eq.~(6) forms a fused expert $\bar{E}_T=\sum_{i=1}^{T}\epsilon_i E_i$, and Eq.~(7) applies it once as $O_{\mathrm{rem}}=\bar{E}_T(h)$~\citep{hidellava-ucit}.
Read as an operator on hidden states, linear fusion implies $\bar{E}_T(h)=\sum_i \epsilon_i E_i(h)=\sum_i \epsilon_i B_i(A_i h)$---the \emph{delta} family rather than the weight-level product $(\sum_i B_i)(\sum_i A_i)$.
We therefore view that \texttt{delta\_sum} reading as closer to the symbolic reading of Eqs.~(6)--(7), whereas the released code implements the separate $A/B$ sum product in Eq.~\eqref{eq:hide_ab_sum}.
Neither reading is usable on FLEX without further correction: magnitude and cross-expert interference both grow with $N$.

\paragraph{Remain-fusion ablation on FLEX.}
Holding experts, backbone, and decoding fixed, we switch to soft top routing ($\tau{=}28$) and vary only the remain fuse (Table~\ref{tab:hide_remain_fuse}).
Unnormalized \texttt{ab\_sum} reaches only $\mathcal{M}{=}13.34$.
The failure is primarily \emph{scale}, not missing experts: \texttt{delta\_sum} adds $N$ raw updates with no $1/N$ factor, so remain updates grow with sequence length; \texttt{ab\_sum} introduces $N^2$ unnormalized cross terms that become severe at $N{=}34$.
Dividing by $N^2$ (\texttt{ab\_mean}) rescales the official product to $\mathcal{M}{=}38.94$ but \emph{retains every cross term} $B_iA_j$ ($i{\neq}j$): incompatible task adaptations can still interfere inside the fused LoRA.
Rescaling alone is therefore insufficient.
We adopt the equal-weight history mix
\begin{equation}
  \Delta_{\mathrm{delta\_mean}}(x)
  =
  \frac{1}{N}\sum_{i=1}^{N} B_i(A_i x) ,
  \label{eq:hide_delta_mean}
\end{equation}
which averages only per-expert updates and excludes cross-expert products, reaching $\mathcal{M}{=}44.39$.
Together with soft top routing at $\tau{=}28$, this is the ``(fixed)'' HiDe Base protocol reported in the main experiment table.

\begin{table}[H]
\centering
\small
\begin{tabular}{llcr}
\toprule
Remain fuse & Top & Ref. & $\mathcal{M}$ \\
\midrule
\texttt{delta\_sum} & hard (top-1) & Eq.~\eqref{eq:hide_delta_sum} & $7.96$ \\
\texttt{ab\_sum} & soft ($\tau{=}28$) & Eq.~\eqref{eq:hide_ab_sum} & $13.34$ \\
\texttt{ab\_mean} & soft ($\tau{=}28$) & Eq.~\eqref{eq:hide_ab_sum}$/N^{2}$ & $38.94$ \\
\texttt{delta\_mean} & soft ($\tau{=}28$) & Eq.~\eqref{eq:hide_delta_mean} & $\mathbf{44.39}$ \\
\bottomrule
\end{tabular}
\caption{FLEX HiDe-LLaVA Base with frozen experts.
Row~1: our initial port (\texttt{delta\_sum}+hard top-1).
Rows~2--4: soft top ($\tau{=}28$), varying remain fuse only.
Row~4: the ``(fixed)'' protocol in the main experiment table.}
\label{tab:hide_remain_fuse}
\end{table}

\paragraph{Direct test of top-routing sensitivity.}
To measure how much generation depends on the top router under this fixed protocol, we evaluate on UCIT~\citep{hidellava-ucit} with the same \texttt{delta\_mean} remain fuse and soft top ($\tau{=}28$), but force every sample to expert~$0$.
Strict matching drops from $99.73\%$ to $16.67\%$, yet $\mathcal{M}$ falls only from $61.32$ to $61.11$ ($-0.21$; Table~\ref{tab:hide_force0}).
Per-task changes stay within about one point; ImageNet-R is unchanged because expert~$0$ is that task's ground-truth expert.

\begin{table}[H]
\centering
\small
\begin{tabular}{lrrr}
\toprule
Task & Normal & Force expert 0 & $\Delta$ \\
\midrule
ImageNet-R & 90.43 & 90.43 & $+0.00$ \\
ArxivQA & 93.70 & 94.03 & $+0.33$ \\
Vizcap & 46.52 & 45.63 & $-0.89$ \\
IconQA & 51.13 & 50.20 & $-0.93$ \\
CLEVR & 30.30 & 30.47 & $+0.17$ \\
Flickr30k & 55.83 & 55.92 & $+0.09$ \\
\midrule
MacroScore & 61.32 & 61.11 & $-0.21$ \\
\bottomrule
\end{tabular}
\caption{HiDe-LLaVA (\texttt{delta\_mean}, soft top $\tau{=}28$) on UCIT: normal routing vs.\ forced expert~$0$.}
\label{tab:hide_force0}
\end{table}

\paragraph{Conclusion.}
Unnormalized remain fusion collapses on FLEX-length sequences; under \texttt{delta\_mean}, even severe top-routing mismatch barely moves $\mathcal{M}$.
The shared remain mixture---not top-layer identification---is the dominant bottleneck.
This explains why stronger CIL routers help HiDe little in the main table, and why frameworks that propagate task-aware mixing beyond the top layer (e.g., PureLoRA) retain more headroom once identification improves.
\FloatBarrier

\subsection{Why DISCO Needs a Frozen Multimodal Projector}
\label{sec:supp_disco_projector}

The original DISCO training protocol jointly updates the task LoRA and the
multimodal projector~\citep{disco}. Although task knowledge is stored in
separate LoRAs, all tasks still share the same projector. Over the long task
sequence of FLEX, the projector is repeatedly updated by each new task.
Consequently, an early LoRA receives visual features that increasingly differ
from those seen when it was trained, even though its own parameters remain
unchanged. This shared representation drift leads to catastrophic forgetting.

We compare DISCO with a frozen versus trainable projector over the 34-task FLEX
sequence. As shown in Table~\ref{tab:disco_projector}, training the projector
reduces MacroScore from 50.32 to 47.07, a drop of 3.25 points, and degrades 25
of the 34 tasks. The forgetting is concentrated toward the beginning of the
sequence: the first 10 tasks lose 5.73 points on average, whereas the last four
lose only 1.35 points.

\begin{table}[H]
\centering
\small
\setlength{\tabcolsep}{4pt}
\renewcommand{\arraystretch}{0.85}
\setlength{\abovecaptionskip}{3pt}
\setlength{\belowcaptionskip}{0pt}
\begin{tabular}{lrrr}
\toprule
Projector & MacroScore & First 10 $\Delta$ & Last 4 $\Delta$ \\
\midrule
Frozen & 50.32 & -- & -- \\
Trainable & 47.07 & $-5.73$ & $-1.35$ \\
\bottomrule
\end{tabular}
\caption{Effect of training the multimodal projector in DISCO. Task-level
changes are measured relative to the frozen-projector setting.}
\label{tab:disco_projector}
\end{table}

Several early tasks are affected particularly strongly. The first two ImageNet
subtasks drop by 7.47 and 7.86 points, IconQA by 8.87 points, and CLEVR by
20.57 points. In contrast, the final task, DCL\_Fin, changes by only $+0.07$
points and is effectively unaffected. Earlier tasks undergo more subsequent
projector updates and therefore suffer more forgetting, while tasks near the
end of the sequence are largely preserved. This trend further indicates that
the degradation comes from continual drift of the shared projector.

Therefore, the multimodal projector cannot be continually fine-tuned along the
task sequence and should remain frozen throughout training. Based on this
result, we use a frozen projector for DISCO on FLEX, and use a "(fixed)" tag.
\FloatBarrier

\subsection{Why SAME Fits Soft Task-as-Class Routing Despite a Fixed Expert Pool}
\label{sec:supp_same_alignment}

The main text classifies SAME as a fixed expert-pool method and notes that
fixed pools generally lack task--expert one-to-one correspondence. SAME is
nonetheless evaluated with the same task-indexed metrics $\mathcal{R}$ and
$\mathcal{G}$ and admits the same plug-in CIL routers as progressive-LoRA
frameworks because its training protocol induces effective
task$\leftrightarrow$expert alignment.

In Prism, expert $k$ is the designated slot for task $k$ ($M=T$; total MoE
rank $34\times 8=272$ on FLEX). While learning task $t$, the outer router
favors a one-hot assignment on expert $t$, and adaptive freezing soft-masks
experts whose activation score
$\mathrm{normalize}(\mathrm{utilization})-\mathrm{normalize}(\mathrm{importance})$
falls below $\tau_{\mathrm{score}}=0.1$, except the current expert
$\min(t,M-1)$, which is never masked. Updates thus concentrate on the
task-aligned slot, and CLIP prototype routing at test time applies soft
weights over the same indexed pool---so $\arg\max_k\alpha_k$ is the same
soft task-as-class decision as in progressive LoRAs.

Table~\ref{tab:same_task_expert_alignment} shows near-saturated Base routing
on UCIT and TriGap ($\mathcal{R}=99.76\%$, $96.07\%$), consistent with
Figure~1, while Table~\ref{tab:same_ucit_diag} confirms per-task top-1
concentration on UCIT (captioning tasks share residual soft mass but keep
$\mathcal{R}\ge98.9\%$). On FLEX the alignment remains defined but
unsaturated ($\mathcal{R}=75.30\%$, $\mathcal{G}=53.65\%$),
leaving room for CIL routers to improve MacroScore. SAME is
therefore a fixed MoE pool whose training-time slot binding makes soft
task-as-class MCIL routing applicable without a separate task-to-expert map.

\begin{table}[H]
\centering
\small
\begin{tabular}{lcccc}
\toprule
Benchmark & Tasks & $\mathcal{M}$ & $\mathcal{G}$ & $\mathcal{R}$ \\
\midrule
UCIT & 6 & 71.84 & 88.35 & 99.76 \\
TriGap & 10 & 45.67 & 91.80 & 96.07 \\
FLEX & 34 & 50.16 & 53.65 & 75.30 \\
\bottomrule
\end{tabular}
\caption{SAME Base routing on UCIT, TriGap, and FLEX. Short
horizons saturate; FLEX retains a larger practical--Oracle gap.}
\label{tab:same_task_expert_alignment}
\end{table}

\begin{table}[H]
\centering
\small
\begin{tabular}{lccc}
\toprule
Task & Dataset & $\mathcal{R}$ & Diagonal $\mathcal{G}$ \\
\midrule
0 & ImageNet-R & 100.00 & 99.95 \\
1 & ArxivQA & 100.00 & 99.47 \\
2 & Vizcap & 98.93 & 68.21 \\
3 & IconQA & 99.90 & 95.61 \\
4 & CLEVR & 100.00 & 97.84 \\
5 & Flickr30k & 99.73 & 69.05 \\
\midrule
Overall / Macro & -- & 99.76 & 88.35 \\
\bottomrule
\end{tabular}
\caption{UCIT per-task SAME Base matching. Diagonal $\mathcal{G}$ is the mean
soft weight on the ground-truth expert index.}
\label{tab:same_ucit_diag}
\end{table}

Although SAME fits soft task-as-class routing, it is substantially slower
than progressive single-adapter baselines on long sequences. The reason is
direct: at each training step SAME updates \emph{multiple} LoRA experts in
the fixed MoE pool, whereas PureLoRA, DISCO, and HiDe primarily optimize
the current-task adapter. As the benchmark grows, more experts enter the
pool and remain eligible for soft routing and (partial) gradient updates,
so wall-clock training time per task increases with sequence length.
Inference inherits a related overhead: soft mixing over a large expert pool
is more expensive than activating a single LoRA.

Table~\ref{tab:same_cost} reports indicative wall-clock measurements under
matched GPU counts. On a 35-task run with 8 GPUs, SAME training takes
about $26.8$\,h ($\sim$47\,min/task) versus $7.3$\,h ($\sim$13\,min/task)
for HiDe ($\sim$3.7$\times$). On single-GPU accelerated evaluation over
ImageNet200 ($\sim$3000 samples), SAME costs about $1230$\,ms/sample versus
$\sim$450\,ms/sample for PureLoRA, DISCO, and HiDe ($\sim$2.7$\times$).

\begin{table}[H]
\centering
\small
\begin{tabular}{llccc}
\toprule
Setting & Method & Hardware & Cost & Rel.\ \\
\midrule
Train  & SAME & 8 GPU & 26.8\,h (47\,min/task) & 3.7$\times$ \\
Train  & HiDe & 8 GPU & 7.3\,h (13\,min/task) & 1.0$\times$ \\
Infer  & SAME & 1 GPU & $\sim$1230\,ms/sample & 2.7$\times$ \\
Infer  & P\&D\&H & 1 GPU & $\sim$450\,ms/sample & 1.0$\times$ \\
\bottomrule
\end{tabular}
\caption{Indicative training and inference cost of SAME versus progressive
LoRA baselines. Relative factors use the fastest method in each block as
$1.0\times$.}
\label{tab:same_cost}
\end{table}
\FloatBarrier

\subsection{Why PureLoRA performs best}
\label{sec:supp_why_purelora_best}

On the formal 34-task \revised{FLEX}, PureLoRA's Base router attains the
highest MacroScore ($54.60$), ahead of DISCO ($50.32$), SAME ($50.16$), and
HiDe-LLaVA ($44.39$).
We attribute this ranking primarily to two routing hyperparameters rather
than to a larger expert capacity: Softmax sharpness and image--text fusion.
Table~\ref{tab:supp_purelora_routing_hparams} isolates each factor with
fixed experts and prototypes.

\paragraph{Soft matching with a sharper Softmax.}
PureLoRA and SAME both use soft prototype routing
$\alpha(\mathbf{x})=\mathrm{softmax}(\tau\mathbf{s}(\mathbf{x}))$ with
logit scale $\tau=28$ (Eq.~\ref{eq:supp_purelora_softmax}).
DISCO instead applies $\mathrm{softmax}(\mathbf{s}/T)$ with
$T{=}0.05$, which is algebraically equivalent to $\tau{=}20$.
Holding PureLoRA fixed and only changing
$\tau{:}\,28\!\rightarrow\!20$ lowers MacroScore from $54.60$ to $52.74$,
moving PureLoRA closer to DISCO's Base ($50.32$).
A sharper soft router therefore explains a non-trivial fraction of
PureLoRA's Base advantage over DISCO.

\paragraph{Balanced image--text fusion.}
PureLoRA fuses modalities with equal weights
$\lambda{=}0.5$ (image) and $1{-}\lambda{=}0.5$ (text)
(Eq.~\ref{eq:supp_purelora_score}).
SAME's Base defaults to a text-heavy mix $\lambda{=}0.2$
(image{:}text $=0.2{:}0.8$), while DISCO is text-only ($\lambda{=}0$).
Holding PureLoRA fixed and only shifting its fusion to the SAME mix
$\lambda{:}\,0.5\!\rightarrow\!0.2$ (no retraining) lowers MacroScore
from $54.60$ to $52.72$ and strict matching from $83.77\%$ to
$79.16\%$, moving PureLoRA closer to SAME's Base ($50.16$) while
remaining above it.
Balanced multimodal fusion therefore also contributes to PureLoRA's
Base advantage over text-heavy or text-only routers.

\begin{table}[H]
\centering
\small
\begin{tabular}{llccc}
\toprule
Setting & Change & $\mathcal{M}$ & $\mathcal{G}$ & $\mathcal{R}$ \\
\midrule
PureLoRA Base & $\tau{=}28$, $\lambda{=}0.5$ & $54.60$ & $60.81$ & $83.77$ \\
PureLoRA $\tau{=}20$ & fixed experts; $\tau{\downarrow}$ & $52.74$ & $49.71$ & $83.77$ \\
PureLoRA $\lambda{=}0.2$ & fixed experts; $\lambda{\downarrow}$ & $52.72$ & $55.93$ & $79.16$ \\
\midrule
SAME Base & $\tau{=}28$, $\lambda{=}0.2$ & $50.16$ & $53.65$ & $75.30$ \\
\bottomrule
\end{tabular}
\caption{Routing-hyperparameter ablations on the formal 34-task
\revised{FLEX} with fixed experts.
Softmax scale $\tau$ and image weight $\lambda$ are varied one at a time.}
\label{tab:supp_purelora_routing_hparams}
\end{table}

Overall, PureLoRA's stronger Base score is consistent with a sharper soft
router ($\tau{=}28$) and a balanced multimodal fusion ($\lambda{=}0.5$).
Lowering $\tau$ to~$20$ or $\lambda$ to~$0.2$ each costs roughly
$1.9$ MacroScore points relative to the default Base ($54.60$), so both
routing choices contribute comparably to PureLoRA's lead.
Matching either knob toward a competing method moves PureLoRA's score
toward that method, indicating that the ranking among Base routers is
sensitive to these routing choices.

\subsection{MCIL Methods Transferable to MCIT Routing}
\label{sec:supp_mcil_transfer}

This subsection continues the main-text reformulation of progressive-LoRA
routing as soft task-as-class MCIL. We detail the four plug-in routers used in
our experiments and clarify which CIL/MCIL mechanisms transfer cleanly.

\paragraph{Transfer protocol.}
Each router replaces only the first-stage task scorer of a host MCIT
framework. CLIP image and text encoders remain frozen; already-trained LoRA
experts, mixture layers, and decoding stay fixed. For every completed task we
extract routing features from that task's training split, update the
task-as-class model, and at inference convert the resulting scores into soft
LoRA weights (or optional hard top-1) under the host framework's mixture rule.
HC, HC-SOINN, and RanPAC maintain separate visual/textual branches and fuse
them with the host modality weights~$\lambda$ (main paper). DISCO's default
HC/HC-SOINN/RanPAC variants are text-only. DDAS instead concatenates the two
CLIP vectors and uses a single autoencoder bank.

\paragraph{HC.}
For each task and modality we buffer up to $10{,}000$ L2-normalized CLIP
features and compress them into $K{=}100$ hierarchical prototypes
(agglomerative clustering with cosine affinity; large buffers first reduce to
$\sim\!1{,}000$ MiniBatchKMeans microcenters). The task score is the maximum
cosine similarity to that task's prototypes, followed by modality fusion and
softmax (PureLoRA/SAME/HiDe use logit scale~$28$; DISCO uses
temperature~$0.05$). No router parameters are trained by backpropagation.

\paragraph{HC-SOINN.}
Each modality builds a prototype graph per task: features are first
compressed to $K_{\mathrm{init}}{=}500$ nodes and then lightly refined by a
SOINN-style update (edge aging and low-degree pruning). The score for task~$t$
combines nearest-class-mean and nearest-node distances,
$d=\alpha d_{\mathrm{NCM}}+(1-\alpha)d_{\mathrm{sub}}$ with $\alpha{=}0.5$,
and uses $\mathrm{logit}_t=-d$. Dual-branch logits are fused and softened with
temperature~$0.03$.

\paragraph{RanPAC.}
Features are L2-normalized, mapped by a fixed random projection of width
$M{=}1024$ with ReLU, and classified by a closed-form ridge regressor
(ridge penalty~$1$) accumulated online as $G\!\leftarrow\!G+h^\top h$ and
$Q\!\leftarrow\!Q+h^\top y$. Inference uses the resulting linear logits,
modality fusion, and temperature~$0.03$. Only the ridge solution is updated;
the projection is never trained.

\paragraph{DDAS.}
The routing vector is $\mathbf{z}^{\mathrm{cat}}=[f_v(\mathbf{v});f_q(\mathbf{q})]$.
Each task owns a one-hidden-layer autoencoder (code dimension~$128$) trained
for $400$ Adam steps on that task's features while freezing older
autoencoders. The score is
$s_k=-\kappa\,\mathrm{MSE}(\mathbf{z}^{\mathrm{cat}},A_k(\mathbf{z}^{\mathrm{cat}}))$
with $\kappa{=}10^{5}$, followed by softmax at temperature~$0.2$. In our
pipeline, DDAS banks are typically fit offline and injected into host
checkpoints so that expert weights remain untouched.

\paragraph{Which methods transfer.}
A CIL/MCIL method is directly compatible when it (i)~treats each incremental
unit as one class that can be identified with an entire MCIT task,
(ii)~emits comparable scores over the expanding historical set, and
(iii)~can be driven by frozen multimodal features without updating LoRA
experts or the generator. Prototype, random-projection, and reconstruction
routers of this form---including HC, HC-SOINN, RanPAC, and DDAS---fit the
protocol above. Methods that require multiple semantic classes inside each
incremental step, explicit intra-task class boundaries, inter-class relation
graphs, or per-class few-shot episodes do not transfer without redefining
their supervision to the coarser task-as-class label. Approaches that couple
classifier learning to backbone updates or that demand generative replay of
expert training data are likewise unsuitable as drop-in routers here, because
they would violate the fixed-expert evaluation. In principle any multimodal
incremental scorer meeting~(i)--(iii) may replace the first-stage router; our
four instantiations are representative rather than exhaustive.

\subsection{Task-Level Routing--Performance Relations}

\begin{figure*}[!t]
\centering
\setlength{\abovecaptionskip}{3pt}
\setlength{\belowcaptionskip}{0pt}
\includegraphics[width=\textwidth]{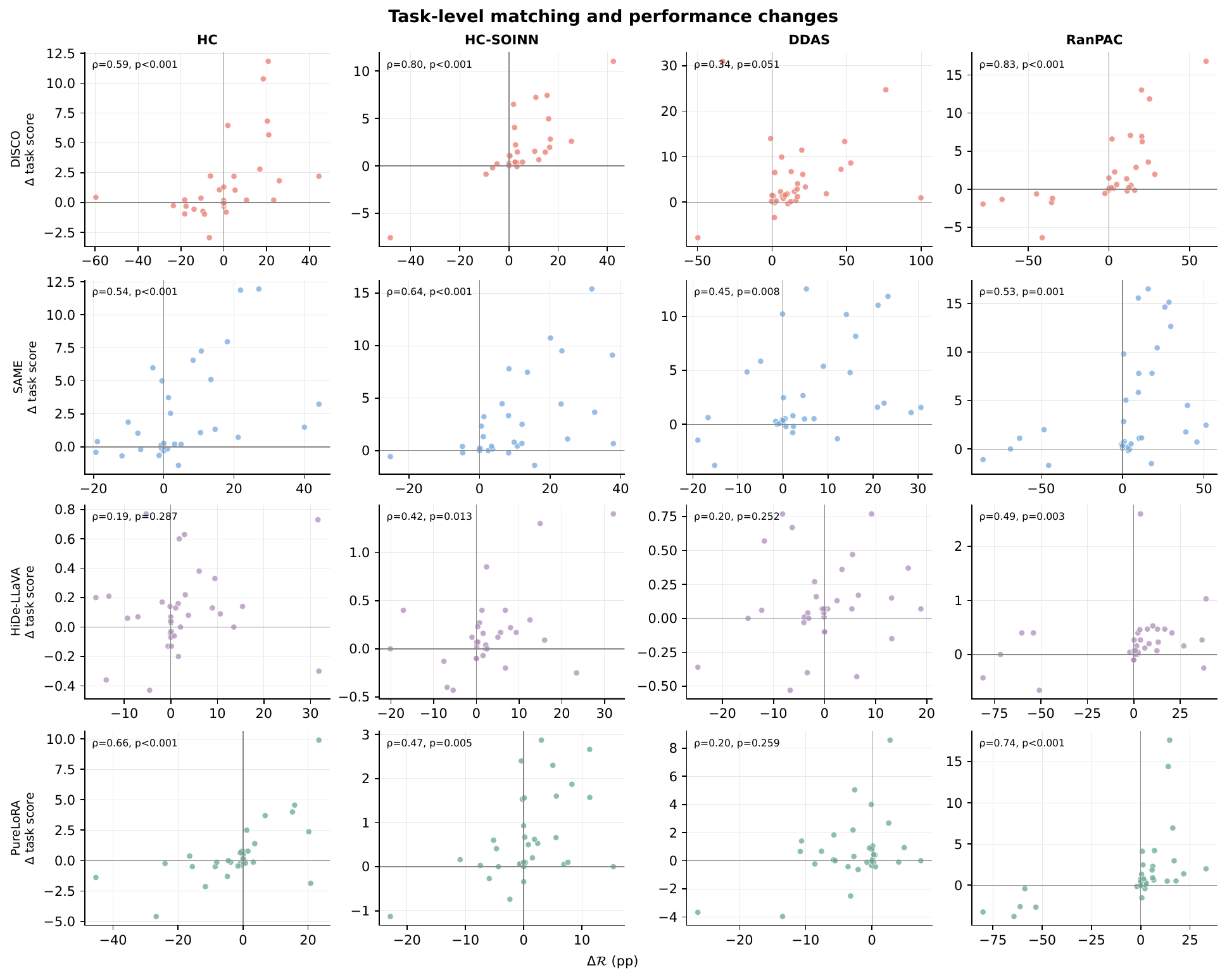}
\caption{Task-level correlations between MacroScore changes and changes in strict matching.}
\label{fig:supp_e4_correlations_match}
\end{figure*}

Aggregate routing improvements correlate with downstream performance, but
task-level effects are heterogeneous. Figure~\ref{fig:supp_e4_correlations_match}
reports the complete task-level correlation views in separate panels: most
method--router rows show a positive association between MacroScore gain and
routing improvement, yet the per-task Spearman correlations remain widely
spread, so a cleaner router does not guarantee a proportional gain on every
task. Figure~\ref{fig:supp_e4_confusion} shows the PureLoRA routing confusion
matrix, where residual errors concentrate among semantically related
neighbors (e.g., ImageNet splits, VQA/knowledge pairs, and document/OCR
tasks) rather than dispersing uniformly; these structured confusions
motivate the six controlled expert-transfer pairs reported in the main paper.

\twocolumn[{
{\captionsetup{type=figure}
\ContinuedFloat
\centering
\setlength{\abovecaptionskip}{2pt}
\setlength{\belowcaptionskip}{0pt}
\includegraphics[width=0.94\textwidth]{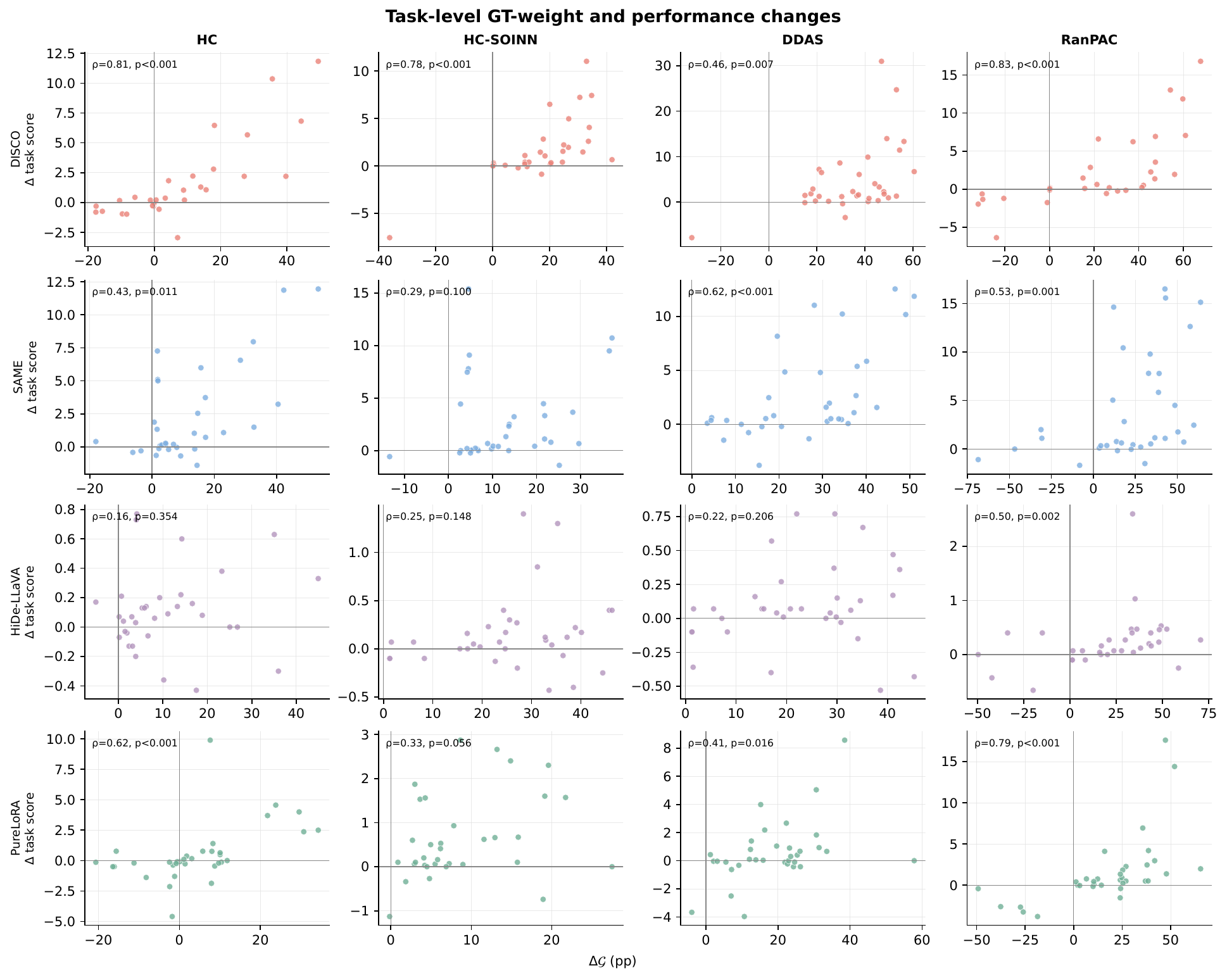}
\caption{Task-level correlations between MacroScore changes and changes in ground-truth LoRA weight.}
\label{fig:supp_e4_correlations_gt}
}
}]

\begin{figure}[!t]
\centering
\setlength{\abovecaptionskip}{2pt}
\setlength{\belowcaptionskip}{0pt}
\includegraphics[width=\columnwidth]{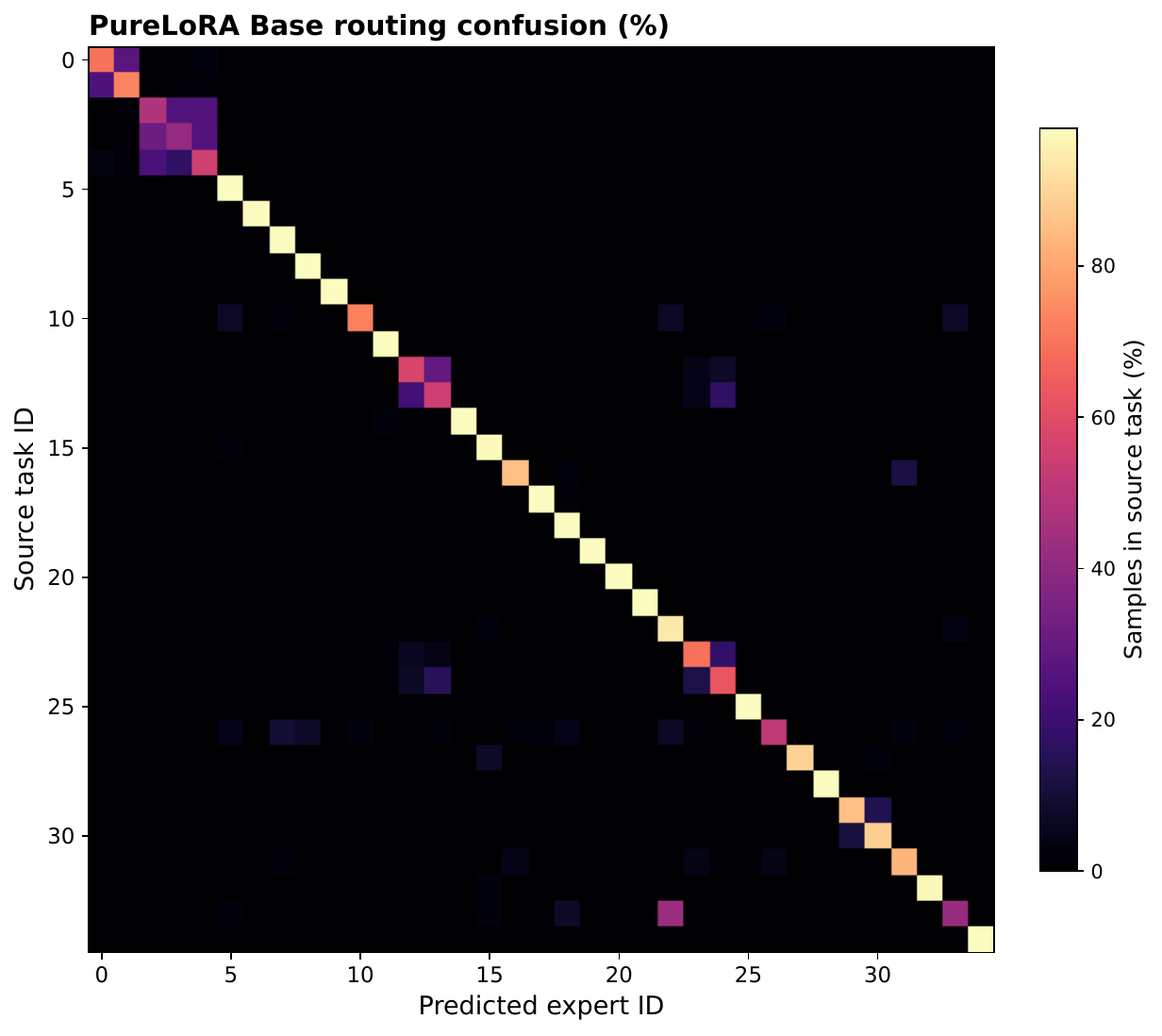}
\caption{PureLoRA task--expert confusion matrix on \revised{FLEX}.}
\label{fig:supp_e4_confusion}
\end{figure}

\subsection{\revised{Complete Task-Level FLEX Results}}
\label{sec:supp_complete_results}

{\color{black}
The compact main-paper table is the abbreviated view: each configuration has
four unweighted task-type means (classification, captioning, multiple choice,
and short-answer VQA) plus the three overall metrics
$\mathcal{M}$, $\mathcal{G}$, and $\mathcal{R}$, for seven reported values.
Table~\ref{tab:supp_main_results_34_tasks} is the detailed view: it expands
those category means into all 34 individual task scores and retains the same
three overall metrics, for 37 reported values per configuration. The four
continued panels preserve the FLEX task order. All 20 configurations are
complete, and SAME+HC-SOINN is included in the means and best/second-best
rankings. Every row uses the pinned FLEX summaries
shared with the main paper, including the official HiDe-LLaVA Base
\texttt{07-25-03-37} run~\citep{coin-moelora,hidellava-ucit,disco,same-trigap,hc-hcsoinn,ranpac,DDAS}.
}

\twocolumn[
\begin{minipage}{\textwidth}
\captionsetup{type=table}
\centering
\small
\setlength{\tabcolsep}{2pt}
\renewcommand{\arraystretch}{0.9}
\begin{tabular}{@{}ll*{9}{r}@{}}
\toprule
MCIT & Router & \rotatebox{60}{IN-1} & \rotatebox{60}{IN-2} & \rotatebox{60}{IN-3} & \rotatebox{60}{IN-4} & \rotatebox{60}{IN-5} & \rotatebox{60}{VizCap} & \rotatebox{60}{Flickr} & \rotatebox{60}{Arxiv} & \rotatebox{60}{Icon} \\
\midrule
DISCO & Base & 28.37 & 42.23 & 52.07 & 51.87 & 49.70 & 46.44 & 56.70 & 94.03 & 54.57 \\
DISCO & HC & 28.10 & 41.90 & 51.77 & 51.70 & 49.60 & 46.39 & 56.66 & \underline{94.20} & \underline{66.40} \\
DISCO & HC-SOINN & \underline{28.63} & \underline{42.53} & \underline{52.17} & \underline{51.93} & \underline{49.87} & \underline{46.45} & 56.70 & \textbf{94.43} & 62.00 \\
DISCO & DDAS & \textbf{59.33} & \textbf{66.93} & \textbf{53.90} & \textbf{59.07} & \textbf{58.30} & \textbf{60.40} & \textbf{57.64} & 93.90 & 66.00 \\
DISCO & RanPAC & 28.57 & 42.33 & 52.07 & \underline{51.93} & 49.80 & 46.34 & \underline{56.76} & 94.13 & \textbf{66.43} \\
\midrule
SAME & Base & 51.73 & 51.13 & 51.03 & 44.97 & 41.10 & 46.52 & 56.88 & 92.07 & 45.97 \\
SAME & HC & 56.83 & 53.00 & 56.03 & 52.23 & 42.43 & 46.64 & 56.94 & \textbf{92.27} & \underline{57.93} \\
SAME & HC-SOINN & \underline{59.53} & 55.57 & \textbf{58.50} & \textbf{60.37} & 50.20 & 46.65 & \underline{56.97} & \underline{92.23} & 55.47 \\
SAME & DDAS & 57.57 & \underline{63.67} & 47.23 & 53.13 & \textbf{52.13} & \textbf{56.74} & \textbf{57.32} & 91.30 & 57.83 \\
SAME & RanPAC & \textbf{67.30} & \textbf{67.63} & \underline{56.07} & \underline{59.60} & \underline{51.53} & \underline{56.31} & \textbf{57.32} & 91.90 & \textbf{58.60} \\
\midrule
HiDe-LLaVA & Base & 81.40 & 65.23 & 47.70 & 55.00 & 48.10 & 38.85 & \underline{52.77} & \underline{78.83} & 37.17 \\
HiDe-LLaVA & HC & \underline{81.53} & \underline{65.37} & 48.43 & 55.77 & 47.97 & \textbf{38.88} & 52.73 & 78.63 & 37.50 \\
HiDe-LLaVA & HC-SOINN & 81.27 & 65.20 & \underline{48.60} & 55.07 & 48.03 & 38.86 & 52.75 & 78.63 & 37.47 \\
HiDe-LLaVA & DDAS & 81.37 & \textbf{65.73} & 48.33 & \underline{56.03} & \underline{49.00} & 38.86 & \textbf{52.81} & \textbf{78.90} & \underline{37.53} \\
HiDe-LLaVA & RanPAC & \textbf{81.57} & \textbf{65.73} & \textbf{49.60} & \textbf{57.80} & \textbf{49.47} & \underline{38.87} & 52.74 & \underline{78.83} & \textbf{37.70} \\
\midrule
PureLoRA & Base & 54.03 & 58.53 & 58.47 & 58.87 & 55.30 & 56.50 & 57.53 & 94.10 & 64.67 \\
PureLoRA & HC & \underline{63.93} & 53.93 & 57.17 & 56.73 & 54.93 & 56.42 & 57.44 & 93.90 & \textbf{67.17} \\
PureLoRA & HC-SOINN & 56.90 & 57.40 & \textbf{60.00} & \underline{59.47} & 57.17 & 58.06 & \textbf{57.59} & \textbf{94.30} & 63.93 \\
PureLoRA & DDAS & 59.07 & \underline{67.10} & 54.50 & 58.93 & \textbf{57.97} & \underline{60.49} & \underline{57.56} & 93.77 & 66.50 \\
PureLoRA & RanPAC & \textbf{71.63} & \textbf{72.93} & \underline{59.10} & \textbf{59.77} & \underline{57.57} & \textbf{60.61} & 57.38 & \underline{94.23} & \underline{67.13} \\
\bottomrule
\end{tabular}
\caption{Detailed task-specific performance and routing quality on the 34-task FLEX benchmark. The four panels report all 34 task scores followed by $\mathcal{M}$, $\mathcal{G}$, and $\mathcal{R}$. Best and second-best results within each MCIT block are bold and underlined. All 20 configurations are complete.}
\label{tab:supp_main_results_34_tasks}
\end{minipage}
\par\medskip
\begin{minipage}{\textwidth}
\captionsetup{type=table}
\ContinuedFloat
\centering
\small
\setlength{\tabcolsep}{2pt}
\renewcommand{\arraystretch}{0.9}
\begin{tabular}{@{}ll*{9}{r}@{}}
\toprule
MCIT & Router & \rotatebox{60}{SQA} & \rotatebox{60}{PMCVQA} & \rotatebox{60}{AI2D} & \rotatebox{60}{TQA} & \rotatebox{60}{DCL\_Fin} & \rotatebox{60}{CLEVR} & \rotatebox{60}{TextVQA} & \rotatebox{60}{GQA} & \rotatebox{60}{VQAv2} \\
\midrule
DISCO & Base & \underline{82.73} & 41.07 & 73.39 & 60.49 & 85.80 & 54.67 & 54.23 & 58.73 & \underline{65.00} \\
DISCO & HC & \textbf{83.77} & 40.27 & 73.60 & 59.93 & 88.60 & 61.13 & 53.50 & \underline{59.80} & \textbf{65.37} \\
DISCO & HC-SOINN & 75.17 & \underline{41.47} & \textbf{74.84} & 62.03 & \underline{88.63} & \underline{61.17} & \textbf{54.43} & 59.07 & 64.93 \\
DISCO & DDAS & 74.90 & 41.30 & \underline{74.64} & \underline{62.08} & \textbf{88.67} & \underline{61.17} & \underline{54.40} & \textbf{60.10} & 64.63 \\
DISCO & RanPAC & 76.37 & \textbf{41.70} & 72.77 & \textbf{62.44} & \textbf{88.67} & \textbf{61.27} & \textbf{54.43} & \textbf{60.10} & 64.77 \\
\midrule
SAME & Base & 77.87 & \underline{41.33} & \underline{71.73} & 57.72 & 87.83 & 56.83 & \textbf{53.40} & 60.10 & \underline{65.17} \\
SAME & HC & 81.10 & 41.03 & 71.31 & \underline{59.21} & 88.10 & \textbf{56.97} & 53.27 & 61.13 & 64.97 \\
SAME & HC-SOINN & \underline{81.53} & \underline{41.33} & 71.52 & 58.39 & 88.03 & 56.83 & \textbf{53.40} & \textbf{61.43} & \textbf{65.60} \\
SAME & DDAS & 79.83 & \underline{41.33} & \textbf{72.35} & 58.80 & \textbf{88.20} & \underline{56.93} & 53.20 & 60.37 & 63.83 \\
SAME & RanPAC & \textbf{82.37} & \textbf{42.10} & \underline{71.73} & \textbf{60.18} & \underline{88.17} & \underline{56.93} & \underline{53.37} & \underline{61.20} & 63.67 \\
\midrule
HiDe-LLaVA & Base & \underline{73.30} & 35.40 & 67.98 & \textbf{53.10} & \underline{63.63} & \textbf{22.50} & 49.63 & 56.93 & 63.03 \\
HiDe-LLaVA & HC & \underline{73.30} & 35.27 & \underline{68.19} & 52.80 & \textbf{63.67} & \underline{22.43} & 49.57 & \underline{57.13} & 63.10 \\
HiDe-LLaVA & HC-SOINN & 73.17 & 35.57 & \textbf{68.40} & 52.90 & \textbf{63.67} & \textbf{22.50} & 49.67 & 57.10 & 63.10 \\
HiDe-LLaVA & DDAS & \textbf{73.37} & \underline{35.67} & 67.98 & \underline{52.95} & 63.53 & 22.40 & \textbf{49.80} & \textbf{57.67} & \textbf{63.33} \\
HiDe-LLaVA & RanPAC & 73.23 & \textbf{35.70} & 67.78 & 52.85 & 63.53 & 22.40 & \underline{49.73} & 56.97 & \underline{63.27} \\
\midrule
PureLoRA & Base & 81.13 & 40.90 & \textbf{75.47} & 61.78 & 88.87 & \textbf{61.57} & 54.13 & 59.00 & 65.30 \\
PureLoRA & HC & \underline{83.50} & \textbf{41.67} & 73.60 & 61.57 & \textbf{89.03} & 61.30 & 53.63 & 58.87 & \textbf{65.67} \\
PureLoRA & HC-SOINN & 82.70 & \underline{41.43} & \textbf{75.47} & 61.88 & \underline{88.97} & 61.23 & 54.20 & 58.73 & \underline{65.33} \\
PureLoRA & DDAS & 80.70 & 41.00 & 74.84 & \underline{62.08} & 88.83 & \underline{61.53} & \underline{54.53} & \textbf{59.93} & 64.87 \\
PureLoRA & RanPAC & \textbf{84.10} & 40.90 & \underline{75.05} & \textbf{63.16} & 88.83 & \textbf{61.57} & \textbf{54.63} & \underline{59.50} & 63.77 \\
\bottomrule
\end{tabular}
\caption{Detailed task-specific performance and routing quality on the 34-task FLEX benchmark (continued).}
\end{minipage}
]

\twocolumn[
\begin{minipage}{\textwidth}
\captionsetup{type=table}
\ContinuedFloat
\centering
\small
\setlength{\tabcolsep}{2pt}
\renewcommand{\arraystretch}{0.9}
\begin{tabular}{@{}ll*{9}{r}@{}}
\toprule
MCIT & Router & \rotatebox{60}{OCRVQA} & \rotatebox{60}{DocVQA} & \rotatebox{60}{ChartQA} & \rotatebox{60}{InfoVQA} & \rotatebox{60}{Roadside} & \rotatebox{60}{ChemVQA} & \rotatebox{60}{FloodNet} & \rotatebox{60}{AOKVQA} & \rotatebox{60}{OKVQA} \\
\midrule
DISCO & Base & \textbf{64.63} & 29.97 & 14.56 & 39.07 & 8.80 & 37.70 & 74.41 & 55.36 & 63.36 \\
DISCO & HC & 64.33 & 29.02 & 14.76 & 38.10 & \underline{11.00} & \underline{39.00} & \underline{81.23} & 55.58 & 63.10 \\
DISCO & HC-SOINN & \underline{64.43} & \underline{30.19} & 16.52 & \textbf{39.47} & 10.27 & 37.77 & 73.54 & \underline{56.46} & \textbf{64.42} \\
DISCO & DDAS & 61.23 & 30.08 & \underline{17.88} & \underline{39.39} & \textbf{11.07} & \textbf{39.17} & 78.45 & \textbf{57.68} & \underline{64.16} \\
DISCO & RanPAC & 64.07 & \textbf{30.48} & \textbf{18.12} & 39.31 & \textbf{11.07} & \textbf{39.17} & \textbf{81.34} & 55.22 & 61.61 \\
\midrule
SAME & Base & 63.60 & 28.06 & 15.04 & 38.89 & 10.80 & 32.27 & 73.27 & 54.71 & 62.70 \\
SAME & HC & \textbf{64.00} & 28.78 & \underline{16.12} & 38.73 & 10.93 & 32.23 & \textbf{81.23} & 54.90 & 62.01 \\
SAME & HC-SOINN & 63.03 & 29.17 & 15.84 & 39.31 & 11.03 & \underline{32.30} & 75.78 & 55.39 & \textbf{63.10} \\
SAME & DDAS & 62.13 & \underline{29.62} & 15.56 & \underline{39.39} & \underline{11.33} & \textbf{32.63} & 78.07 & \textbf{56.29} & \underline{62.77} \\
SAME & RanPAC & \underline{63.80} & \textbf{29.82} & \textbf{16.20} & \textbf{39.42} & \textbf{11.43} & \textbf{32.63} & \underline{81.07} & \underline{55.43} & 61.02 \\
\midrule
HiDe-LLaVA & Base & 61.23 & 23.25 & 10.04 & 36.20 & \underline{7.63} & \underline{13.73} & 46.54 & 54.11 & \underline{61.91} \\
HiDe-LLaVA & HC & \underline{61.40} & 23.34 & 10.04 & 36.28 & \textbf{7.70} & 13.70 & \textbf{46.70} & \underline{54.17} & 61.55 \\
HiDe-LLaVA & HC-SOINN & 61.20 & 23.31 & \underline{10.08} & 36.27 & 7.60 & \underline{13.73} & \underline{46.59} & 54.08 & \underline{61.91} \\
HiDe-LLaVA & DDAS & 60.93 & \textbf{23.44} & \textbf{10.16} & \textbf{36.33} & 7.53 & \textbf{13.80} & \textbf{46.70} & 54.14 & 61.28 \\
HiDe-LLaVA & RanPAC & \textbf{61.47} & \underline{23.40} & 10.04 & \underline{36.31} & 7.53 & \textbf{13.80} & \textbf{46.70} & \textbf{54.39} & \textbf{61.94} \\
\midrule
PureLoRA & Base & \underline{64.27} & 30.18 & 16.76 & 39.40 & 10.37 & 38.30 & 81.07 & 56.91 & 64.26 \\
PureLoRA & HC & 64.13 & 30.05 & 18.16 & 38.90 & 11.13 & 38.40 & \textbf{81.56} & 56.67 & 62.87 \\
PureLoRA & HC-SOINN & \textbf{64.33} & \underline{30.23} & \underline{18.36} & \textbf{40.02} & \textbf{11.30} & 38.40 & 81.07 & 57.32 & \underline{64.42} \\
PureLoRA & DDAS & 60.60 & 30.08 & 17.80 & 39.29 & \underline{11.17} & \textbf{38.73} & 78.56 & \underline{57.58} & \textbf{64.92} \\
PureLoRA & RanPAC & 63.87 & \textbf{30.70} & \textbf{18.60} & \underline{39.64} & 11.13 & \underline{38.70} & \underline{81.51} & \textbf{58.90} & 60.46 \\
\bottomrule
\end{tabular}
\caption{Detailed task-specific performance and routing quality on the 34-task FLEX benchmark (continued).}
\end{minipage}
\par\medskip
\begin{minipage}{\textwidth}
\captionsetup{type=table}
\ContinuedFloat
\centering
\small
\setlength{\tabcolsep}{2pt}
\renewcommand{\arraystretch}{0.9}
\begin{tabular}{@{}ll*{10}{r}@{}}
\toprule
MCIT & Router & \rotatebox{60}{ArtVQA} & \rotatebox{60}{MathVista} & \rotatebox{60}{PathVQA} & \rotatebox{60}{PlantVQA} & \rotatebox{60}{SLAKE} & \rotatebox{60}{VQA-RAD} & \rotatebox{60}{ACL-OCR} & $\mathcal{M}$ & $\mathcal{G}$ & $\mathcal{R}$ \\
\midrule
DISCO & Base & 18.67 & 25.80 & 38.27 & 50.83 & 56.27 & 42.57 & 38.39 & 50.32 & 43.90 & 63.95 \\
DISCO & HC & 15.73 & \underline{28.00} & 40.10 & 56.50 & \underline{66.63} & \textbf{44.79} & 38.83 & 51.75 & 51.18 & 64.58 \\
DISCO & HC-SOINN & 22.73 & \textbf{28.40} & 49.30 & 55.80 & 63.50 & \textbf{44.79} & \underline{39.05} & 51.86 & 59.63 & 66.87 \\
DISCO & DDAS & \underline{25.37} & 27.00 & \underline{51.60} & \underline{56.90} & 66.16 & \underline{44.35} & \textbf{39.70} & \textbf{54.93} & \textbf{78.86} & \textbf{80.21} \\
DISCO & RanPAC & \textbf{25.73} & 24.60 & \textbf{55.07} & \textbf{57.07} & \textbf{69.30} & 41.24 & 36.44 & \underline{52.37} & \underline{63.90} & \underline{70.37} \\
\midrule
SAME & Base & 18.13 & 28.00 & 38.30 & 50.80 & 58.65 & 40.13 & \textbf{38.83} & 50.16 & 53.65 & 75.30 \\
SAME & HC & 20.67 & 26.60 & \underline{50.17} & \textbf{54.53} & \underline{65.21} & \textbf{46.12} & 38.18 & 52.41 & 65.20 & 81.03 \\
SAME & HC-SOINN & 22.60 & 26.60 & 49.03 & 53.13 & 61.88 & \underline{43.46} & \underline{38.61} & 52.76 & 66.19 & \underline{84.97} \\
SAME & DDAS & \underline{23.50} & \underline{28.80} & 48.47 & 53.27 & 63.50 & 42.79 & \underline{38.61} & \underline{52.90} & \textbf{79.79} & 81.33 \\
SAME & RanPAC & \textbf{23.97} & \textbf{30.00} & \textbf{53.43} & \underline{53.63} & \textbf{66.44} & 41.24 & 37.74 & \textbf{54.21} & \underline{74.76} & \textbf{86.40} \\
\midrule
HiDe-LLaVA & Base & 9.23 & 26.20 & 27.67 & 26.30 & 39.45 & \underline{37.47} & \underline{37.74} & 44.39 & 56.59 & 79.69 \\
HiDe-LLaVA & HC & \textbf{9.37} & \underline{26.80} & \textbf{28.30} & \textbf{26.43} & \underline{39.83} & \textbf{37.69} & 37.31 & 44.50 & 67.58 & 82.44 \\
HiDe-LLaVA & HC-SOINN & 9.23 & \textbf{27.00} & 27.73 & \underline{26.40} & 39.73 & \underline{37.47} & \textbf{37.96} & 44.48 & 64.60 & \underline{82.56} \\
HiDe-LLaVA & DDAS & \underline{9.30} & 25.80 & 27.77 & 26.37 & \textbf{40.02} & \underline{37.47} & 37.31 & \underline{44.52} & \textbf{79.46} & 80.21 \\
HiDe-LLaVA & RanPAC & \underline{9.30} & 26.60 & \underline{28.20} & 26.37 & \underline{39.83} & 37.25 & 37.31 & \textbf{44.65} & \underline{76.63} & \textbf{86.27} \\
\midrule
PureLoRA & Base & 24.43 & 26.40 & 51.93 & 56.80 & 63.88 & \underline{45.68} & \textbf{39.48} & 54.60 & 60.81 & 83.77 \\
PureLoRA & HC & 25.07 & \textbf{30.40} & \underline{55.63} & \textbf{57.57} & \underline{68.44} & 45.23 & \textbf{39.48} & 55.12 & 65.13 & 81.44 \\
PureLoRA & HC-SOINN & 25.10 & \underline{28.80} & 54.23 & \underline{57.30} & 66.54 & \textbf{46.34} & \textbf{39.48} & \underline{55.24} & 69.60 & \underline{83.92} \\
PureLoRA & DDAS & \underline{25.33} & 27.80 & 51.70 & 56.70 & 66.06 & \underline{45.68} & \textbf{39.48} & 55.17 & \textbf{79.79} & 81.33 \\
PureLoRA & RanPAC & \textbf{25.77} & 23.80 & \textbf{56.13} & \textbf{57.57} & \textbf{70.82} & 43.02 & \underline{36.23} & \textbf{56.14} & \underline{78.32} & \textbf{87.72} \\
\bottomrule
\end{tabular}
\caption{Detailed task-specific performance and routing quality on the 34-task FLEX benchmark (continued).}
\end{minipage}
]

\FloatBarrier

\section{Additional Experiments}
\label{sec:supp_additional_experiments}

\subsection{CLIP Representation Extraction}

\begin{figure*}[!t]
\centering
\includegraphics[width=\textwidth]{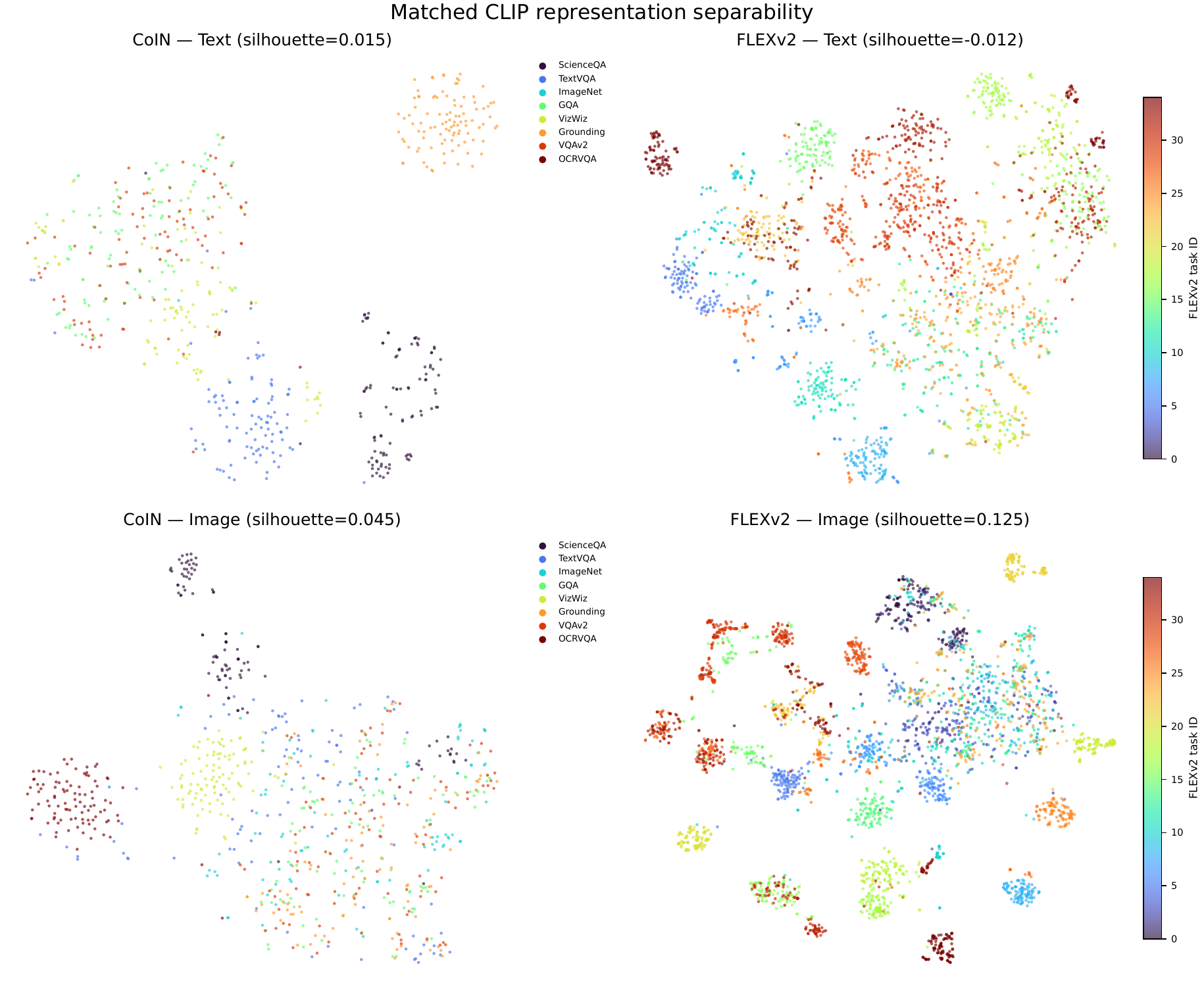}
\caption{CLIP image and deduplicated text representations used by the routing analysis.}
\label{fig:a1_representation}
\end{figure*}

We extract paired CLIP ViT-L/14-336 image and text features for up to 500
samples per task from CoIN~\citep{coin-moelora} and \revised{FLEX}. Text vectors are deduplicated within each
task, all features are $\ell_2$-normalized, and cosine t-SNE~\citep{t-sne} uses a fixed seed
and perplexity 30. Figure~\ref{fig:a1_representation} is included only as an
implementation-level visualization of the extracted routing features.

\FloatBarrier

\subsection{Router Efficiency}

The added routing state occupies at most 52.77 MiB, or 4.98\% of the
corresponding full checkpoint. Router-only latency is 61.65--76.41 ms per
sample, and end-to-end latency changes by only $-2.45\%$ to $+0.97\%$
relative to Base. The transferred classifiers therefore add little deployment
memory and no substantial inference penalty under the controlled protocol.

\begin{table}[H]
\centering
\small
\setlength{\tabcolsep}{2pt}
\renewcommand{\arraystretch}{0.78}
\setlength{\abovecaptionskip}{3pt}
\setlength{\belowcaptionskip}{0pt}
\begin{tabular}{lrr}
\toprule
Router & Extra (MiB) & Extra (\%) \\
\midrule
Base & 0.00 & 0.00\% \\
HC & 0.00--20.13 & 0.00--1.48\% \\
HC-SOINN & 21.15--48.91 & 1.48--4.63\% \\
RanPAC & 7.28--14.55 & 0.51--1.42\% \\
DDAS & 52.77 & 3.55--4.98\% \\
\bottomrule
\end{tabular}

\vspace{2pt}
\begin{tabular}{lrrr}
\toprule
Router & Router (ms) & End-to-end (ms) & E2E change \\
\midrule
Base & 63.12 $\pm$ 0.28 & 258.65 $\pm$ 3.88 & +0.00\% \\
HC & 76.41 $\pm$ 2.00 & 261.15 $\pm$ 0.72 & +0.97\% \\
HC-SOINN & 67.57 $\pm$ 4.73 & 257.11 $\pm$ 1.40 & -0.60\% \\
RanPAC & 61.65 $\pm$ 3.44 & 252.32 $\pm$ 0.56 & -2.45\% \\
DDAS & 65.07 $\pm$ 4.49 & 255.42 $\pm$ 3.08 & -1.25\% \\
\bottomrule
\end{tabular}
\caption{Controlled memory and inference overhead of CIL routers. Extra memory is reported in MiB and as a percentage of the corresponding full model checkpoint. Latency is reported as mean $\pm$ standard deviation.}
\label{tab:router_efficiency}
\end{table}

\begin{figure}[H]
\centering
\includegraphics[width=0.9\columnwidth]{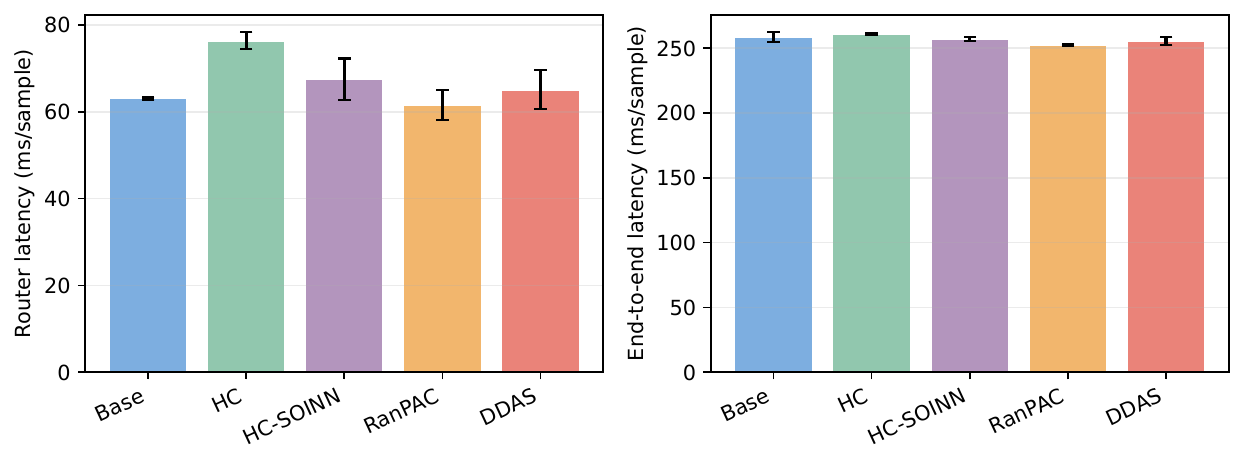}
\caption{Controlled router-only and end-to-end inference latency.}
\label{fig:a4_latency}
\end{figure}
\FloatBarrier

\subsection{Transfer to UCIT and TriGap}

Without retuning router identity, the selected CIL routers improve routing
quality on both benchmarks. DISCO+DDAS improves MacroScore by 2.37 on UCIT
and 1.14 on TriGap~\citep{same-trigap}; the remaining integrations maintain or improve
MacroScore while consistently increasing strict matching.
On TriGap, HiDe-LLaVA+RanPAC changes MacroScore only from $37.39$ to $37.36$
($-0.03$) while raising $\mathcal{G}$ and $\mathcal{R}$.
Thus, transferring MCIL classifiers to MCIT routing is not specific to
\revised{FLEX}.

\begin{table}[H]
\centering
\small
\begin{minipage}[t]{\columnwidth}
\centering
\textbf{UCIT}\\[2pt]
\begin{tabular}{llrrr}
\toprule
Method & Router & $\mathcal{M}$ & $\mathcal{G}$ & $\mathcal{R}$ \\
\midrule
DISCO & Base & 66.10 & 79.68 & 82.93 \\
DISCO & DDAS & 68.47 & 99.78 & 99.80 \\
SAME & Base & 71.84 & 88.35 & 99.76 \\
SAME & RanPAC & 74.36 & 99.79 & 99.98 \\
HiDe-LLaVA & Base & 65.37 & 99.77 & 99.77 \\
HiDe-LLaVA & RanPAC & 66.35 & 99.96 & 99.98 \\
PureLoRA & Base & 76.20 & 94.65 & 99.75 \\
PureLoRA & RanPAC & 76.49 & 99.96 & 99.98 \\
\bottomrule
\end{tabular}
\end{minipage}\par\medskip
\begin{minipage}[t]{\columnwidth}
\centering
\textbf{TriGap}\\[2pt]
\begin{tabular}{llrrr}
\toprule
Method & Router & $\mathcal{M}$ & $\mathcal{G}$ & $\mathcal{R}$ \\
\midrule
DISCO & Base & 41.23 & 80.95 & 91.63 \\
DISCO & DDAS & 42.37 & 98.29 & 97.71 \\
SAME & Base & 45.67 & 91.80 & 96.07 \\
SAME & RanPAC & 45.85 & 98.85 & 98.60 \\
HiDe-LLaVA & Base & 37.39 & 95.08 & 98.45 \\
HiDe-LLaVA & RanPAC & 37.36 & 99.67 & 99.67 \\
PureLoRA & Base & 46.19 & 95.23 & 98.78 \\
PureLoRA & RanPAC & 46.24 & 99.68 & 99.67 \\
\bottomrule
\end{tabular}
\end{minipage}
\caption{Cross-benchmark transfer of routers selected on \revised{FLEX}.}
\label{tab:a5_cross_benchmark}
\end{table}

\begin{figure*}[!t]
\centering
\includegraphics[width=0.90\textwidth]{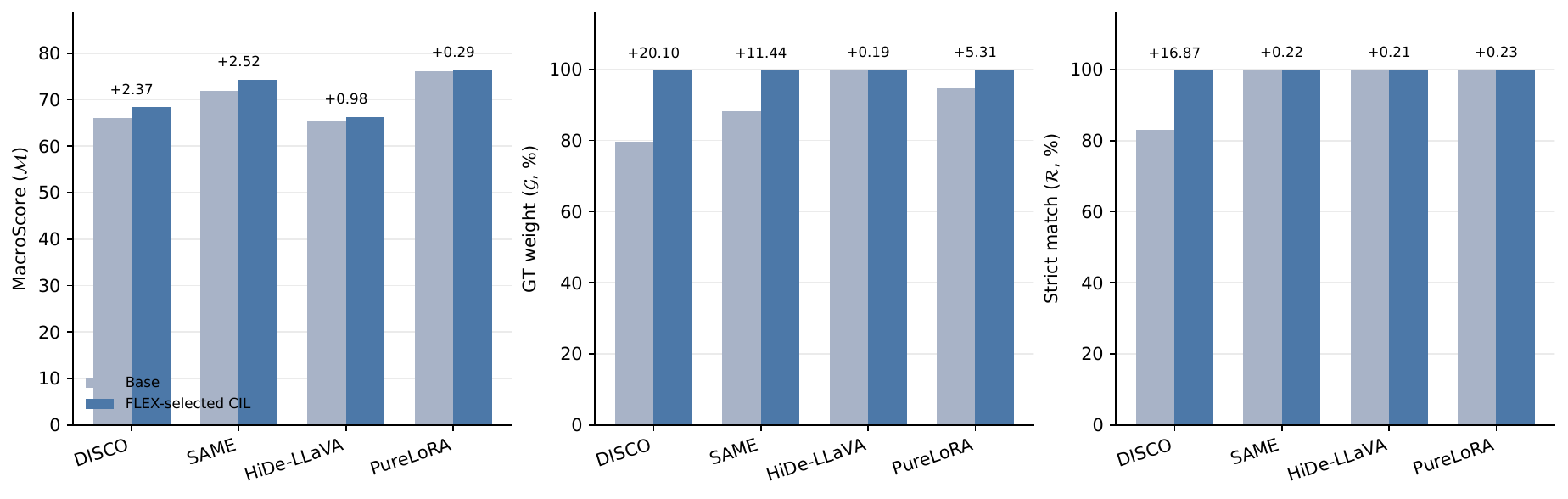}\par\medskip
\includegraphics[width=0.90\textwidth]{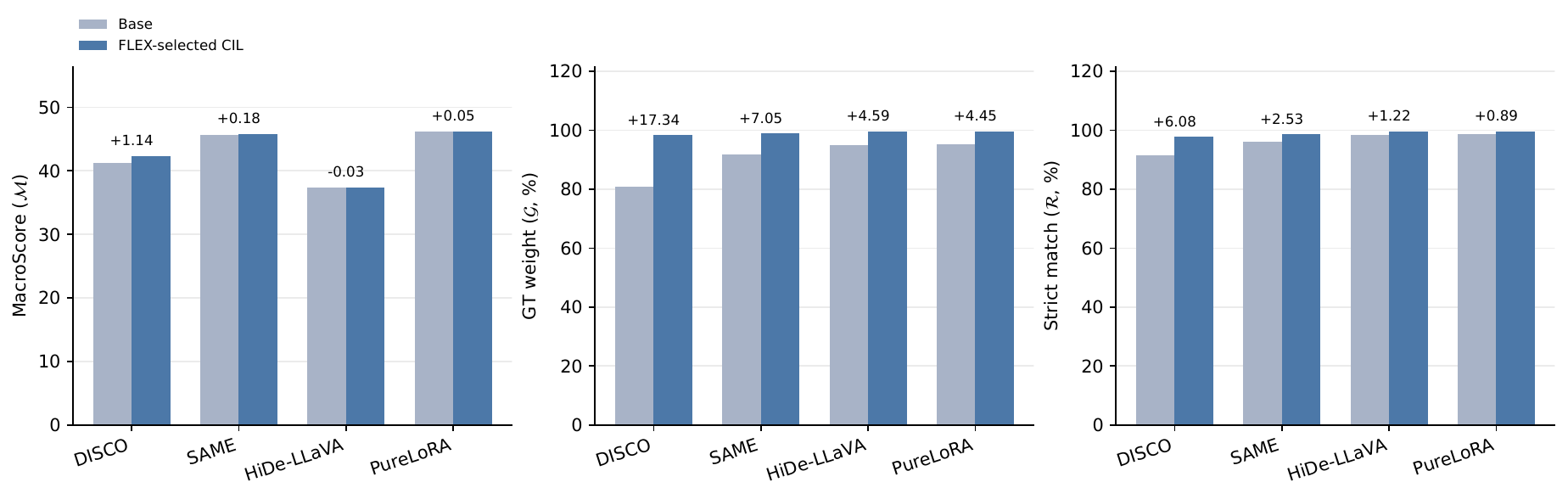}
\caption{Base and CIL routing on UCIT (top) and TriGap (bottom).}
\label{fig:a5_cross_benchmark}
\end{figure*}
\FloatBarrier

\subsection{Seed Robustness}
\label{sec:supp_seed_robustness}

{\color{black}
To assess sensitivity of the main-table conclusions to implementation
randomness, we report two seed studies: (i)~router-initialization variance
under a fixed expert bank, obtained by varying only RanPAC's
random-projection seed; and (ii)~expert-training variance under a fixed task
order and evaluation protocol, obtained by changing the global training seed
of HiDe-LLaVA.
The first study uses PureLoRA+RanPAC; the second evaluates Base, HC, and
RanPAC on HiDe-LLaVA.
We focus on HiDe-LLaVA because it exhibits the smallest MacroScore gains in
the main table and is therefore the configuration most plausibly attributable
to random fluctuation; the training-seed study tests whether those small
gains vanish or reverse under expert retraining.

\subsubsection{Router Projection Seeds}

With experts, replay features, fusion weights, temperature, and test sets
held fixed, we vary only RanPAC's random-projection seed.
PureLoRA+RanPAC attains MacroScore $53.60\pm0.03$ and strict matching
$88.09\pm0.07$.
The small variation confirms that the PureLoRA+RanPAC gains in the main table
are robust to projection initialization.
}

\begin{table}[H]
\centering
\small
\begin{tabular}{rrrr}
\toprule
RanPAC seed & $\mathcal{M}$ & $\mathcal{G}$ & $\mathcal{R}$ \\
\midrule
20260710 & 53.58 & 78.29 & 88.01 \\
20260711 & 53.63 & 78.23 & 88.10 \\
20260712 & 53.59 & 78.34 & 88.15 \\
\bottomrule
\end{tabular}
\caption{PureLoRA+RanPAC over three random-projection seeds.}
\label{tab:a6_seed}
\end{table}

\begin{figure}[H]
\centering
\includegraphics[width=0.9\columnwidth]{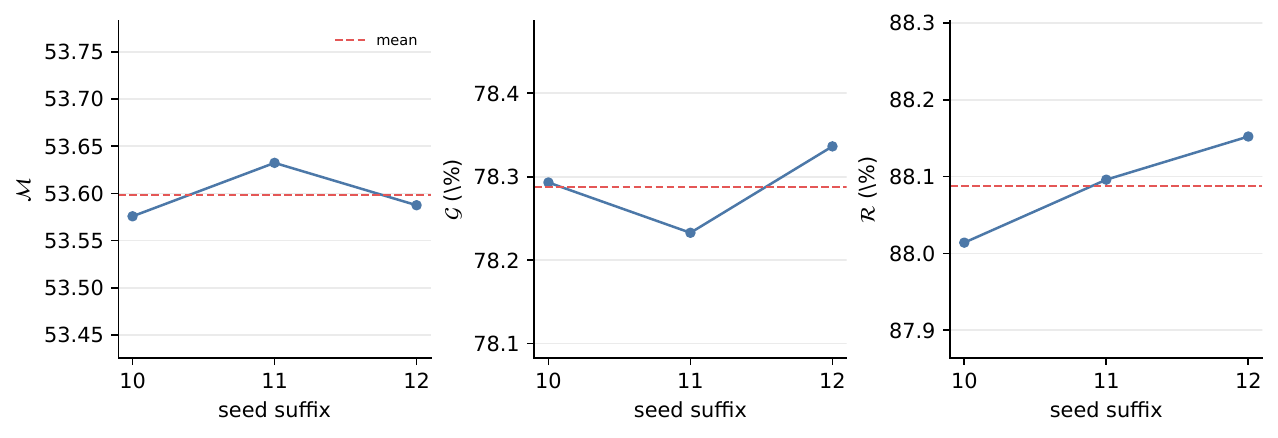}
\caption{Performance variation across RanPAC projection seeds.}
\label{fig:a6_seed}
\end{figure}

\subsubsection{Expert Training Seeds}

{\color{black}
We retrain HiDe-LLaVA-HC on the canonical TFPv3 task order for the main-table
canonical run and two additional global training seeds ($20260729$ and
$20260730$), yielding three training trajectories.
Each trajectory is evaluated under the main-table protocol with Base, HC, and
RanPAC (soft routing, $\tau{=}28$).
The RanPAC router is transplanted from a fixed PureLoRA checkpoint, so router
parameters are identical across trajectories and differences arise only from
the retrained experts.

Results are reported in Table~\ref{tab:hide_seed}.
Across the three trajectories, MacroScore is highly stable to the training
seed: Base $44.38\pm0.02$ (range $0.04$), HC $44.44\pm0.06$ (range $0.12$),
and RanPAC $44.64\pm0.02$ (range $0.03$).
Gains relative to Base keep a consistent sign: HC yields
$\Delta\mathcal{M}\in[+0.03,+0.11]$ (mean $+0.06$), and RanPAC yields
$\Delta\mathcal{M}\in[+0.26,+0.27]$ (mean $+0.27$); the small MacroScore
improvements do not reverse sign across training seeds.
Routing-quality metrics improve more substantially and remain stable:
relative to Base, HC gains about $+11$ points in $\mathcal{G}$ and
$+2.5$ points in $\mathcal{R}$ on average.
Because RanPAC uses a fixed transplanted router, $\mathcal{G}=76.63$ and
$\mathcal{R}=86.27$ are identical across the three trajectories, while
MacroScore varies by only $0.03$ points due to expert differences.

Overall, the main-table pattern for HiDe-LLaVA---improved
$\mathcal{G}/\mathcal{R}$ with a small MacroScore gain---is not an artifact
of a single training seed.
The small $\Delta\mathcal{M}$ remains directionally consistent and stable
under expert retraining, and is therefore difficult to attribute to training
randomness alone.
}

\begin{table}[H]
\centering
\small
\begin{tabular}{llrrr}
\toprule
Training seed & Router & $\mathcal{M}$ & $\mathcal{G}$ & $\mathcal{R}$ \\
\midrule
canonical & Base & 44.39 & 56.59 & 79.69 \\
canonical & HC & 44.50 & 67.58 & 82.44 \\
canonical & RanPAC & 44.65 & 76.63 & 86.27 \\
20260729 & Base & 44.35 & 55.17 & 78.78 \\
20260729 & HC & 44.38 & 66.31 & 80.57 \\
20260729 & RanPAC & 44.62 & 76.63 & 86.27 \\
20260730 & Base & 44.39 & 55.47 & 78.65 \\
20260730 & HC & 44.44 & 66.32 & 81.60 \\
20260730 & RanPAC & 44.66 & 76.63 & 86.27 \\
\midrule
mean$\pm$std & Base & 44.38$\pm$0.02 & 55.74$\pm$0.75 & 79.04$\pm$0.57 \\
mean$\pm$std & HC & 44.44$\pm$0.06 & 66.74$\pm$0.73 & 81.54$\pm$0.94 \\
mean$\pm$std & RanPAC & 44.64$\pm$0.02 & 76.63$\pm$0.00 & 86.27$\pm$0.00 \\
\bottomrule
\end{tabular}
\caption{HiDe-LLaVA Base/HC/RanPAC on FLEX across three training seeds.
RanPAC uses a fixed PureLoRA-transplanted router, so $\mathcal{G}$ and
$\mathcal{R}$ are identical across seeds.}
\label{tab:hide_seed}
\end{table}
\FloatBarrier

\subsection{InternVL Backbone}

{\color{black}Replacing LLaVA with InternVL-Chat-ViT-6B-Vicuna-7B~\citep{internvl} gives a Base MacroScore of 51.14, 60.27\% ground-truth weight, and 83.30\% strict matching. GT one-hot routing reaches 56.60, leaving a 5.46-point routing gap concentrated in the five ImageNet subsets (+29.48 points on average there). All four transferred CIL routers are evaluated with the same trained experts; 4/4 improve MacroScore and ground-truth weight, while 2/4 improve strict matching. RanPAC is best ($\mathcal{M}=53.06$, +1.92 over Base; $\mathcal{R}=87.93\%$), recovering 35.1\% of the Base-to-Oracle gap---comparable to the 39.7\% recovery on LLaVA. DDAS assigns the highest ground-truth weight (79.10\%) but gains less MacroScore (+1.04), and HC-SOINN improves matching (+0.39\%) without a large score lift (+0.70). These results confirm that MCIL-to-MCIT router transfer generalizes beyond LLaVA, with ImageNet-style classification remaining the dominant residual bottleneck.}

\begin{table}[H]
\centering
\small
\setlength{\tabcolsep}{4pt}
\begin{tabular}{lrrrr}
\toprule
Router & $\mathcal{M}$ & $\Delta\mathcal{M}$ & $\mathcal{G}$ (\%) & $\mathcal{R}$ (\%) \\
\midrule
Base & 51.14 & +0.00 & 60.27 & 83.30 \\
+ HC & 52.01 & +0.87 & 64.92 & 80.91 \\
+ HC-SOINN & 51.84 & +0.70 & 69.33 & 83.69 \\
+ RanPAC & 53.06 & +1.92 & 78.07 & 87.93 \\
+ DDAS & 52.18 & +1.04 & 79.10 & 80.80 \\
GT one-hot & 56.60 & +5.46 & 100.00 & 100.00 \\
\bottomrule
\end{tabular}
\caption{\revised{PureLoRA routing with the InternVL backbone on FLEX. All variants share the same trained LoRA experts; only the router differs.}}
\label{tab:internvl_backbone}
\end{table}
\FloatBarrier

\subsection{Qualitative Routing Failures}

The failures span scientific diagrams, fine-grained classification, general
and knowledge VQA, document understanding, and medical VQA. They justify the
strict one-task--one-LoRA matching definition: treating related experts as
correct would conceal errors that remove task-specific capabilities.

\begin{figure*}[!t]
\centering
\includegraphics[width=\textwidth]{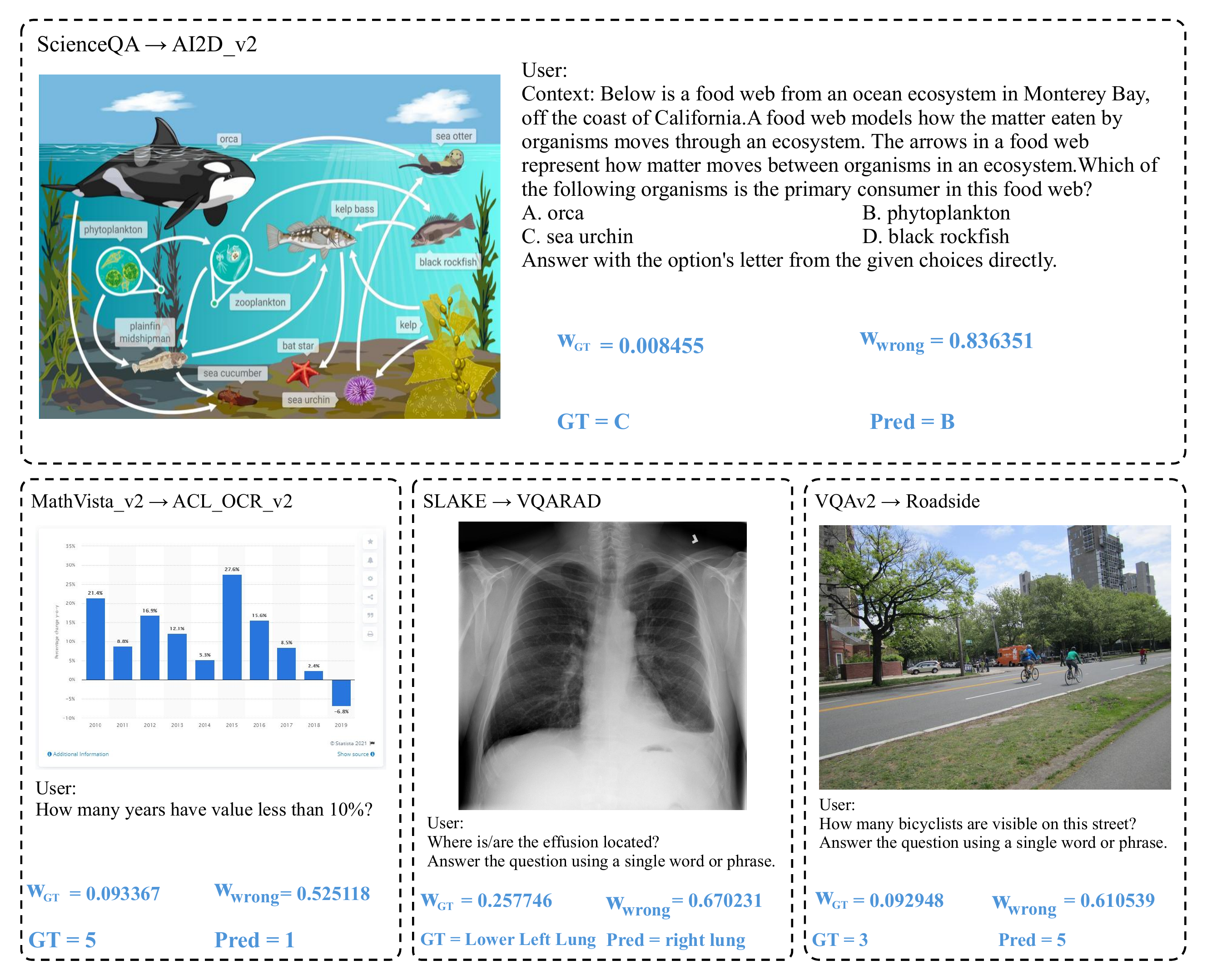}
\caption{Four representative PureLoRA routing failures. Semantically related but task-incorrect experts receive high weight and produce incorrect answers.}
\label{fig:routing_failures}
\end{figure*}
\FloatBarrier

\subsection{FLEX Task Catalog}

\revised{FLEX} is a 34-task multimodal continual instruction-tuning
benchmark designed to weaken textual identity leakage while retaining
substantive visual-domain and knowledge differences among tasks.
We group datasets that share an outer answer interface---image
classification, captioning, multiple choice, or short-answer VQA---and
normalize only the shared trailing answer-format template within each group
when such a template is present.
Question bodies, option layouts, and dataset-native special fields are
retained for fidelity.
ImageNet-1K is split into five class-disjoint 200-class tasks so that the
classification interface alone cannot identify a unique LoRA.
Captioning is represented by Vizcap and Flickr30k; multiple-choice tasks
span scientific figures, education, medicine, diagrams, textbooks, and
finance; the remaining short-answer tasks cover natural scenes, OCR and
documents, charts, traffic, chemistry, art, plant pathology, and clinical
imaging.
For datasets of sufficient size we sample 10{,}000 training and 3{,}000 test
examples (30{,}000 / 3{,}000 per ImageNet subset); smaller sets keep all
available samples, yielding 388{,}856 training and 85{,}037 test examples in
total.
Native evaluation metrics are preserved per dataset.
Table~\ref{tab:a8_suffix_map} lists every continual task with its domain,
answer style, and the task-specific frozen answer-format suffix used as the
Strong fingerprint protocol in Section~\ref{sec:supp_ablation}.
Citations are attached under each dataset name to avoid crowding the main
text.

\begin{table*}[t]
\centering
\footnotesize
\setlength{\tabcolsep}{2.5pt}
\renewcommand{\arraystretch}{1.12}
\setlength{\abovecaptionskip}{4pt}
\setlength{\belowcaptionskip}{0pt}
\begin{tabularx}{\textwidth}{@{}
  r
  l
  >{\raggedright\arraybackslash}p{2.2cm}
  c
  >{\raggedright\arraybackslash}X
  @{}}
\toprule
ID & Dataset & Domain & Style & Frozen suffix \\
\midrule
00 & \flexds{ImageNet200\_1}{Deng_2009} & Natural images & short & Express your answer in a single word or a short, descriptive phrase. \\
01 & \flexds{ImageNet200\_2}{Deng_2009} & Natural images & short & Provide your answer using a single word or a brief phrase. \\
02 & \flexds{ImageNet200\_3}{Deng_2009} & Natural images & short & Describe the content of the image using one word or a concise phrase. \\
03 & \flexds{ImageNet200\_4}{Deng_2009} & Natural images & short & Respond to the question with a single word or a short, descriptive phrase. \\
04 & \flexds{ImageNet200\_5}{Deng_2009} & Natural images & short & Classify the image content using only one word or a brief phrase. \\
05 & \flexds{ArxivQA}{li-etal-2024-multimodal-arxiv} & Sci.\ figures & choice & Select the correct answer from the given choices and respond with the letter of the chosen option. \\
06 & \flexds{Vizcap}{Gurari_2020} & Assistive caption & caption & Provide a brief caption that describes the image. \\
07 & \flexds{IconQA}{lu2021iconqa} & Abstract icons & short & Respond to the question with a single word or a short phrase. \\
07 & \flexds{IconQA}{lu2021iconqa} & Abstract icons & choice & Determine the correct option from the provided choices and reply with its corresponding letter. \\
08 & \flexds{CLEVR}{Johnson_2017} & Synthetic scenes & short & Respond to the question using only one word or a concise phrase. \\
09 & \flexds{Flickr30k}{young-etal-2014-image} & Natural caption & caption & Describe the image in one concise caption. \\
10 & \flexds{ScienceQA}{Lu_2022} & Science education & choice & Pick the correct answer from the listed options and provide the letter of the selected option. \\
11 & \flexds{TextVQA}{Singh_2019} & Scene text & short & Answer the question with a single word or a brief phrase. \\
12 & \flexds{GQA}{Hudson_2019} & Comp.\ VQA & short & Respond with one word or a short phrase. \\
\bottomrule
\end{tabularx}
\caption{\revised{Complete FLEX task catalog (34 continual tasks).
Each row lists domain, answer style, and the task-specific frozen
answer-format suffix used as the Strong fingerprint protocol
(Section~\ref{sec:supp_ablation}; Table~\ref{tab:fingerprint_suffix_diff}).
Mixed-format datasets have separate rows per style (same ID).
Dataset citations appear under each dataset name.}}
\label{tab:a8_suffix_map}
\end{table*}

\begin{table*}[t]
\ContinuedFloat
\centering
\footnotesize
\setlength{\tabcolsep}{2.5pt}
\renewcommand{\arraystretch}{1.12}
\setlength{\abovecaptionskip}{4pt}
\setlength{\belowcaptionskip}{0pt}
\begin{tabularx}{\textwidth}{@{}
  r
  l
  >{\raggedright\arraybackslash}p{2.2cm}
  c
  >{\raggedright\arraybackslash}X
  @{}}
\toprule
ID & Dataset & Domain & Style & Frozen suffix \\
\midrule
13 & \flexds{VQAv2}{Goyal_2017} & Natural VQA & short & Provide your answer in the form of a single word or a concise phrase. \\
14 & \flexds{OCRVQA}{Mishra_2019} & Book-cover OCR & short & Respond to the question with just one word or a brief phrase. \\
15 & \flexds{PMCVQA}{zhang2024pmcvqavisualinstructiontuning} & Biomedical VQA & choice & Identify the correct choice from the options below and respond with the letter of the correct option. \\
16 & \flexds{DocVQA}{Mathew_2021} & Document VQA & short & Answer the question using a single word or a concise phrase. \\
17 & \flexds{ChartQA}{masry-etal-2022-chartqa} & Charts & short & Provide your response using only one word or a short phrase. \\
18 & \flexds{InfographicVQA}{Mathew_2022} & Infographic & short & Respond to the question with a single word or a brief phrase. \\
19 & \flexds{Roadside}{Guan_2026} & Roadside traffic & short & Respond to the question using just one word or a concise phrase. \\
20 & \flexds{ChemVQA}{chemvqa2k} & Chemistry & short & Answer the question with one word or a short phrase. \\
21 & \flexds{FloodNetVQA}{Rahnemoonfar_2021} & Aerial / flood & short & Give your answer as a single word or a concise phrase. \\
22 & \flexds{AI2D}{Kembhavi_2016} & Science diagrams & choice & From the given choices, choose the correct answer and respond with the letter of that choice. \\
23 & \flexds{AOKVQA}{Schwenk_2022} & Knowledge VQA & short & Provide one word or a short phrase as your answer. \\
24 & \flexds{OKVQA}{Marino_2019} & Knowledge VQA & short & Use a single word or a brief phrase for your response. \\
25 & \flexds{ArtVQA}{Garcia_2020} & Artwork & short & Reply using one word or a concise phrase. \\
26 & \flexds{MathVista}{lu2024mathvista} & Math reasoning & short & State the final answer in a single word or a short phrase. \\
26 & \flexds{MathVista}{lu2024mathvista} & Math reasoning & choice & Choose the right answer from the options and respond with its letter. \\
27 & \flexds{PathVQA}{he2020pathvqa30000questionsmedical} & Pathology & short & Give a one-word answer or a brief descriptive phrase. \\
28 & \flexds{PlantVillageVQA}{sakib2026plantexpertvqavisualquestionanswering} & Plant disease & short & Provide the answer as one word or a concise phrase. \\
29 & \flexds{SLAKE}{Liu_2021} & Clinical VQA & short & Reply to the question with a single word or a brief phrase. \\
30 & \flexds{VQARAD}{Lau_2018} & Radiology & short & Use just one word or a short phrase in your answer. \\
31 & \flexds{ACL\_OCR}{Liu_2024} & Document OCR & short & Answer using only a single word or a concise phrase. \\
32 & \flexds{TQA}{Kembhavi_2017} & Textbook science & choice & Select the correct answer from the provided options and reply with the letter associated with it. \\
33 & \flexds{DCL\_Fin}{zhao2025mllm} & Financial charts & short & Respond with a single word or a concise phrase. \\
33 & \flexds{DCL\_Fin}{zhao2025mllm} & Financial charts & choice & From the given choices, select the correct answer and reply with the letter of the chosen option. \\
\bottomrule
\end{tabularx}
\caption{\revised{Complete FLEX task catalog (continued).
Mixed-format rows share an ID.}}
\end{table*}

\paragraph{Frozen suffix protocol.}
Table~\ref{tab:a8_suffix_map} lists the Strong (task-specific) answer-format
suffixes used consistently whenever the fingerprint protocol is Strong.
Only the trailing answer-format instruction is replaced; question bodies and
special fields are unchanged.
The Base$\leftrightarrow$Strong mapping for every task is given in
Table~\ref{tab:fingerprint_suffix_diff} of Section~\ref{sec:supp_ablation}.
\FloatBarrier

\subsection{Ablation Studies}
\label{sec:supp_ablation}

FLEX simultaneously lengthens the continual sequence and weakens
interface-to-identity leakage, so a natural concern is whether its hardness
comes from the larger competing LoRA pool, from reduced textual uniqueness, or
from both.
Although these factors co-occur in benchmark construction---adding confusable
peers under a shared outer answer format necessarily grows $T$---they are
separately manipulable at evaluation time.
We report a factorial fingerprint$\times$sequence-regime table for aesthetics
and within-row fingerprint contrasts
(Table~\ref{tab:flex_2x2_ablation}), and a prefix-matched expert-pool scaling
table that isolates pool size under a fixed evaluation set
(Table~\ref{tab:flex_pool_scaling}).

\paragraph{What we do not ablate.}
Fingerprint reduction is \emph{not} implemented by deleting task-necessary
content such as TextVQA's ``\texttt{Reference OCR token:}'' field, multiple-choice
option blocks, or long question bodies (e.g., ArxivQA).
Removing such fields would change the underlying skill rather than the
identity cue.
Instead, FLEX retains necessary content, normalizes only the shared outer
answer-format template within each interface group when present, and
introduces confusable peers (five ImageNet-200 splits; paired captioning and
multiple-choice datasets) so that an interface no longer maps one-to-one onto
a single LoRA.

\begin{figure}[H]
\centering
\includegraphics[width=0.9\columnwidth]{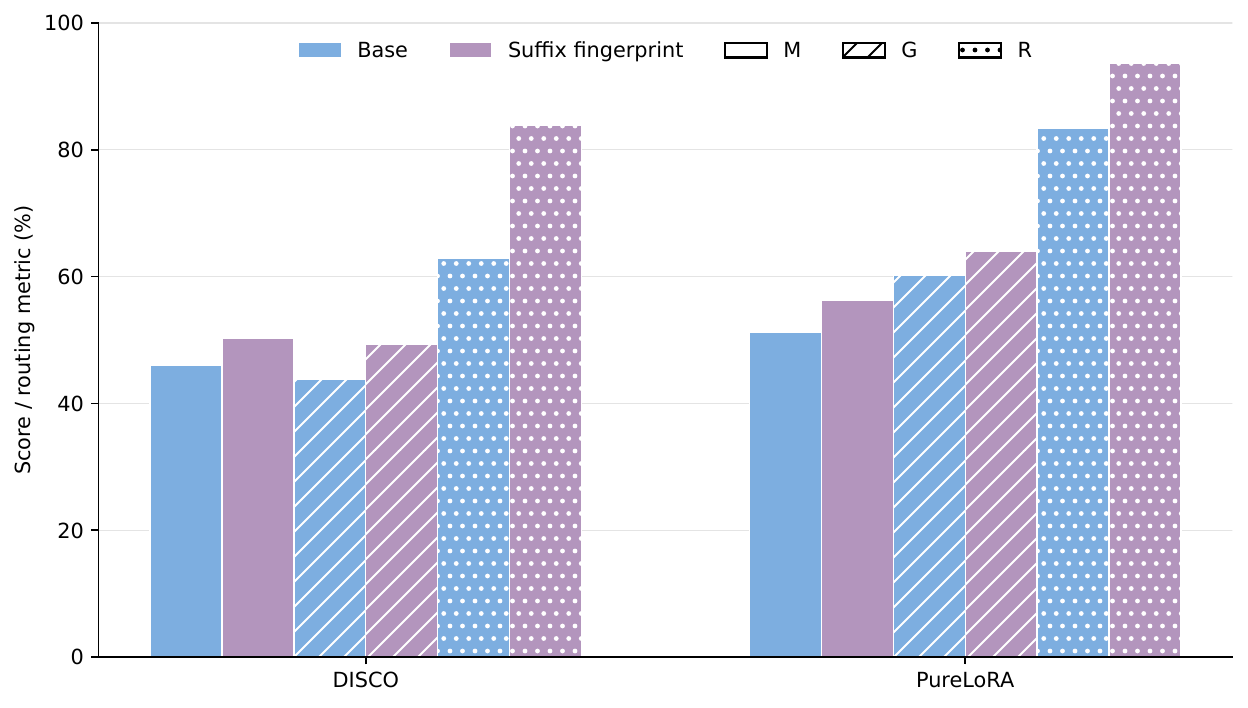}
\caption{Effect of task-specific answer-suffix fingerprints on routing and performance
(Long$\times$Strong vs.\ Long$\times$Weak; see Table~\ref{tab:flex_2x2_ablation}).}
\label{fig:a8_fingerprint}
\end{figure}

\paragraph{Fingerprint protocol.}
\label{sec:supp_fingerprint}
Weak keeps the shared trailing answer-format templates of the base FLEX
release.
Strong replaces only that trailing instruction with a task-specific semantic
paraphrase during both training and evaluation, leaving question bodies and
dataset-native special fields unchanged.
Table~\ref{tab:fingerprint_suffix_diff} lists the Base versus Strong suffix for
every FLEX task; all Strong cells below follow this table (Short and
prefix-matched rows use only the relevant task subset of the mapping).
A dash denotes rows with no shared trailing answer-format suffix in the base
release.
Restoring Strong suffixes isolates the textual factor and recovers easier
routing relative to Weak under the same sequence regime
(Figure~\ref{fig:a8_fingerprint}; Table~\ref{tab:flex_2x2_ablation}).

\begin{table*}[t]
\centering
\footnotesize
\setlength{\tabcolsep}{2.5pt}
\renewcommand{\arraystretch}{1.12}
\setlength{\abovecaptionskip}{4pt}
\setlength{\belowcaptionskip}{0pt}
\begin{tabularx}{\textwidth}{@{}
  r
  l
  >{\raggedright\arraybackslash}X
  >{\raggedright\arraybackslash}X
  @{}}
\toprule
ID & Dataset & Base suffix & Fingerprint suffix \\
\midrule
00 & ImageNet200\_1 & Answer the question using a single word or phrase. & Express your answer in a single word or a short, descriptive phrase. \\
01 & ImageNet200\_2 & Answer the question using a single word or phrase. & Provide your answer using a single word or a brief phrase. \\
02 & ImageNet200\_3 & Answer the question using a single word or phrase. & Describe the content of the image using one word or a concise phrase. \\
03 & ImageNet200\_4 & Answer the question using a single word or phrase. & Respond to the question with a single word or a short, descriptive phrase. \\
04 & ImageNet200\_5 & Answer the question using a single word or phrase. & Classify the image content using only one word or a brief phrase. \\
05 & ArxivQA & Answer with the option's letter from the given choices directly. & Select the correct answer from the given choices and respond with the letter of the chosen option. \\
06 & Vizcap & Generate a brief caption for the image. & Provide a brief caption that describes the image. \\
07 & IconQA (short) & Answer the question using a single word or phrase. & Respond to the question with a single word or a short phrase. \\
07 & IconQA (choice) & Answer with the option's letter from the given choices directly. & Determine the correct option from the provided choices and reply with its corresponding letter. \\
08 & CLEVR & Answer the question using a single word or phrase. & Respond to the question using only one word or a concise phrase. \\
09 & Flickr30k & Generate a brief caption for the image. & Describe the image in one concise caption. \\
10 & ScienceQA & Answer with the option's letter from the given choices directly. & Pick the correct answer from the listed options and provide the letter of the selected option. \\
11 & TextVQA & Answer the question using a single word or phrase. & Answer the question with a single word or a brief phrase. \\
12 & GQA & Answer the question using a single word or phrase. & Respond with one word or a short phrase. \\
13 & VQAv2 & Answer the question using a single word or phrase. & Provide your answer in the form of a single word or a concise phrase. \\
14 & OCRVQA & Answer each question using a single word or phrase. & Respond to the question with just one word or a brief phrase. \\
15 & PMCVQA & Answer with the option's letter. & Identify the correct choice from the options below and respond with the letter of the correct option. \\
16 & DocVQA & -- & Answer the question using a single word or a concise phrase. \\
17 & ChartQA & Answer the question using a single word or phrase. & Provide your response using only one word or a short phrase. \\
18 & InfographicVQA & -- & Respond to the question with a single word or a brief phrase. \\
\bottomrule
\end{tabularx}
\caption{Base versus fingerprint trailing answer-format suffixes on \revised{FLEX}.
All Strong ablations use this mapping (Short and prefix-matched rows use
tasks $0$--$9$ only).
Mixed-format datasets have separate rows per style (same ID).
A dash indicates no shared trailing format template in the base release.}
\label{tab:fingerprint_suffix_diff}
\end{table*}

\begin{table*}[t]
\ContinuedFloat
\centering
\footnotesize
\setlength{\tabcolsep}{2.5pt}
\renewcommand{\arraystretch}{1.12}
\setlength{\abovecaptionskip}{4pt}
\setlength{\belowcaptionskip}{0pt}
\begin{tabularx}{\textwidth}{@{}
  r
  l
  >{\raggedright\arraybackslash}X
  >{\raggedright\arraybackslash}X
  @{}}
\toprule
ID & Dataset & Base suffix & Fingerprint suffix \\
\midrule
19 & Roadside & Answer the question using a single word or phrase. & Respond to the question using just one word or a concise phrase. \\
20 & ChemVQA & Answer the question using a single word or phrase. & Answer the question with one word or a short phrase. \\
21 & FloodNetVQA & Answer the question using a single word or phrase. & Give your answer as a single word or a concise phrase. \\
22 & AI2D & Answer with the option's letter from the given choices directly. & From the given choices, choose the correct answer and respond with the letter of that choice. \\
23 & AOKVQA & Answer the question using a single word or phrase. & Provide one word or a short phrase as your answer. \\
24 & OKVQA & Answer the question using a single word or phrase. & Use a single word or a brief phrase for your response. \\
25 & ArtVQA & Answer the question using a single word or phrase. & Reply using one word or a concise phrase. \\
26 & MathVista (short) & -- & State the final answer in a single word or a short phrase. \\
26 & MathVista (choice) & Answer with the option's letter from the given choices directly. & Choose the right answer from the options and respond with its letter. \\
27 & PathVQA & Answer the question using a single word or phrase. & Give a one-word answer or a brief descriptive phrase. \\
28 & PlantVillageVQA & Answer the question using a single word or phrase. & Provide the answer as one word or a concise phrase. \\
29 & SLAKE & Answer the question using a single word or phrase. & Reply to the question with a single word or a brief phrase. \\
30 & VQARAD & Answer the question using a single word or phrase. & Use just one word or a short phrase in your answer. \\
31 & ACL\_OCR & -- & Answer using only a single word or a concise phrase. \\
32 & TQA & Answer with the option's letter from the given choices directly. & Select the correct answer from the provided options and reply with the letter associated with it. \\
33 & DCL\_Fin (short) & -- & Respond with a single word or a concise phrase. \\
33 & DCL\_Fin (choice) & Answer with the given letter directly, e.g., A, B, C, D. & From the given choices, select the correct answer and reply with the letter of the chosen option. \\
\bottomrule
\end{tabularx}
\caption{Base versus fingerprint trailing answer-format suffixes on \revised{FLEX} (continued).}
\end{table*}

\paragraph{LoRA reuse.}
DISCO and PureLoRA train task-independent LoRA experts and do not update the
multimodal projector.
Reduced-pool cells therefore reuse experts $0,\ldots,T{-}1$ from the final
34-task checkpoint and truncate the router to a $T$-way pool, rather than
retraining.
This reuse keeps expert parameters identical across pool sizes, so observed
differences can be attributed to routing competition instead of retrain
variance.
Strong reduced-pool cells likewise reuse experts from the corresponding
Strong long-sequence checkpoint trained under
Table~\ref{tab:fingerprint_suffix_diff}.
All reduced-pool evaluations use a true $T$-way softmax; they are not macros
over the first $T$ tasks of a 34-way router.
For the $T{=}34$ prefix-matched rows in
Table~\ref{tab:flex_pool_scaling}, we reuse the full 34-way Long runs and
report metrics only on tasks $0$--$9$.

\paragraph{Factorial design.}
Table~\ref{tab:flex_2x2_ablation} reports DISCO and PureLoRA under a
$2{\times}2$ of fingerprint protocol and sequence regime.
Short is an early-sequence slice: tasks $0$--$9$ with a $T{=}10$ expert pool
(ImageNet200$_1$--$_5$, ArxivQA, Vizcap, IconQA, CLEVR, Flickr30k).
Long is the formal 34-task FLEX sequence with a 34-way pool.
Weak keeps the shared base answer-format suffixes; Strong uses the
task-specific paraphrases in Table~\ref{tab:fingerprint_suffix_diff}.
Long$\times$Weak matches the formal FLEX Base runs in the main paper;
Long$\times$Strong is the full-sequence suffix-fingerprint injection.
The primary reading is \emph{within-row}: Short$\times$Weak vs.\
Short$\times$Strong and Long$\times$Weak vs.\ Long$\times$Strong isolate the
textual factor under a fixed evaluation set and pool size.
Cross-row MacroScore comparisons are \emph{not} difficulty contrasts, because
Short and Long evaluate different task sets; the contribution of pool size
under a matched evaluation set is isolated in
Table~\ref{tab:flex_pool_scaling}.
Within each regime, restoring Strong suffixes sharply improves routing:
on Short, DISCO raises $\mathcal{R}$ from $49.69$ to $99.26$ and PureLoRA from
$79.38$ to $99.70$; on Long, both methods likewise gain in
$\mathcal{M}$, $\mathcal{G}$, and $\mathcal{R}$.
Thus superficial answer-format uniqueness alone can drive near-ceiling
matching even when experts are held fixed.

\begin{table}[H]
\centering
\small
\begin{tabular}{llrrr}
\toprule
Method & Setting & $\mathcal{M}$ & $\mathcal{G}$ & $\mathcal{R}$ \\
\midrule
\multicolumn{5}{@{}l@{}}{\emph{Short (tasks $0$--$9$, $T{=}10$), weak fingerprints}} \\
DISCO & Short$\times$Weak & 54.81 & 46.94 & 49.69 \\
PureLoRA & Short$\times$Weak & 63.29 & 62.37 & 79.38 \\
\midrule
\multicolumn{5}{@{}l@{}}{\emph{Short (tasks $0$--$9$, $T{=}10$), strong fingerprints}} \\
DISCO & Short$\times$Strong & 72.77 & 65.94 & 99.26 \\
PureLoRA & Short$\times$Strong & 74.66 & 75.13 & 99.70 \\
\midrule
\multicolumn{5}{@{}l@{}}{\emph{Long (full FLEX, $T{=}34$), weak fingerprints}} \\
DISCO & Long$\times$Weak & 50.32 & 43.90 & 63.95 \\
PureLoRA & Long$\times$Weak & 54.60 & 60.81 & 83.77 \\
\midrule
\multicolumn{5}{@{}l@{}}{\emph{Long (full FLEX, $T{=}34$), strong fingerprints}} \\
DISCO & Long$\times$Strong & 56.66 & 51.00 & 86.36 \\
PureLoRA & Long$\times$Strong & 58.69 & 65.68 & 91.40 \\
\bottomrule
\end{tabular}
\caption{Fingerprint$\times$sequence-regime ablation on \revised{FLEX} with
DISCO and PureLoRA.
Weak uses shared outer answer-format suffixes; Strong uses the task-specific
paraphrases in Table~\ref{tab:fingerprint_suffix_diff}.
Short reuses the first $T{=}10$ independently trained LoRAs from the
corresponding long checkpoint with a truncated $T$-way router.
Read primarily within rows; do not cross-compare Short and Long MacroScore as
matched difficulty.}
\label{tab:flex_2x2_ablation}
\end{table}

\paragraph{Prefix-matched expert-pool scaling.}
Table~\ref{tab:flex_pool_scaling} isolates the contribution of the long
expert pool under a fixed evaluation set.
We always evaluate tasks $0$--$9$ and vary only the competing LoRA count
$T\in\{10,22,34\}$, truncating the same final LoRA bank and router to a true
$T$-way softmax (for $T{=}34$, metrics are aggregated from the full Long runs
on the same prefix).
$T{=}10$ matches the Short cells of Table~\ref{tab:flex_2x2_ablation};
$T{=}34$ matches the early-task regime of formal FLEX under a 34-way pool;
$T{=}22$ is an intermediate midpoint.
The clean pool-size signal is soft routing quality: under both Weak and Strong
protocols, $\mathcal{G}_{0:9}$ decreases monotonically as $T$ grows for DISCO
and PureLoRA, so long-horizon difficulty is not merely an artifact of later,
harder datasets.
Under Weak fingerprints, DISCO matching stays far from saturation
($\mathcal{R}_{0:9}$ in the mid-$40$s across $T$), whereas Strong suffixes keep
$\mathcal{R}_{0:9}$ near ceiling ($\ge\!97.8$) even at $T{=}34$, while still
showing the same $\mathcal{G}$ dilution.
Prefix MacroScore tracks the same direction more mildly for both methods
(Weak and Strong), while PureLoRA's $\mathcal{R}_{0:9}$ remains nearly flat
under Weak and near-saturated under Strong.
Together with Table~\ref{tab:flex_2x2_ablation}, the design attributes FLEX
difficulty to the joint effect of reduced textual uniqueness and a larger
competing expert pool.

\begin{table}[H]
\centering
\small
\begin{tabular}{llrrr}
\toprule
Method & Protocol / $T$ & $\mathcal{M}_{0:9}$ & $\mathcal{G}_{0:9}$ & $\mathcal{R}_{0:9}$ \\
\midrule
DISCO & Weak / 10 & 54.81 & 46.94 & 49.69 \\
DISCO & Weak / 22 & 53.93 & 42.28 & 47.00 \\
DISCO & Weak / 34 & 53.06 & 39.71 & 46.69 \\
DISCO & Strong / 10 & 72.77 & 65.94 & 99.26 \\
DISCO & Strong / 22 & 72.17 & 61.89 & 98.55 \\
DISCO & Strong / 34 & 71.47 & 58.82 & 97.80 \\
\midrule
PureLoRA & Weak / 10 & 63.29 & 62.37 & 79.38 \\
PureLoRA & Weak / 22 & 62.73 & 59.61 & 79.21 \\
PureLoRA & Weak / 34 & 61.96 & 56.61 & 79.04 \\
PureLoRA & Strong / 10 & 74.66 & 75.13 & 99.70 \\
PureLoRA & Strong / 22 & 74.28 & 72.98 & 99.60 \\
PureLoRA & Strong / 34 & 73.52 & 68.11 & 98.39 \\
\bottomrule
\end{tabular}
\caption{Prefix-matched expert-pool scaling on \revised{FLEX}.
All rows evaluate only tasks $0$--$9$ under a true $T$-way router
($T\in\{10,22,34\}$).
Strong rows follow Table~\ref{tab:fingerprint_suffix_diff}.
Reduced-pool rows reuse LoRAs $0,\ldots,T{-}1$ from the corresponding final
FLEX checkpoint; $T{=}34$ aggregates the Long runs on the same prefix.}
\label{tab:flex_pool_scaling}
\end{table}
\FloatBarrier

\subsection{Task Order}
\label{sec:supp_task_order}

This experiment measures the effect of task acquisition order on
\revised{FLEX} performance and routing quality.
DISCO, PureLoRA, and HiDe-LLaVA assign each dataset its own LoRA and train that
LoRA on the corresponding data only, so later tasks do not overwrite earlier
experts through a shared adapter; for these methods, acquisition order does
not rewrite the expert bank.
We therefore evaluate order sensitivity only on SAME, reporting native soft
routing (Base) and the plug-in CIL router (HC).

The canonical \revised{FLEX} order is the fixed list in
Table~\ref{tab:a8_suffix_map} (ID~$0$--$33$).
It is a catalog order, not a ranking by routing difficulty or curriculum
score.
Holding the task set, data splits, backbone, and metrics fixed, we vary only
the presentation order of the 34 dataset names: the canonical catalog order,
plus two full random permutations of those names drawn with seeds $20260730$
and $20260731$.
For each order we train SAME once and evaluate Base and HC on the same
checkpoint.
We report $\mathcal{M}$, $\mathcal{G}$, and $\mathcal{R}$ as in the main text,
with mean and sample standard deviation over the three orders ($n{=}3$).

\begin{table}[H]
\centering
\small
\setlength{\tabcolsep}{4.5pt}
\providecommand{\pmcell}[2]{\begin{tabular}{@{}r@{}}#1\\[-0.12em]{\scriptsize$\pm$#2}\end{tabular}}
\begin{tabular}{llrrr}
\toprule
Order & Router & $\mathcal{M}$ & $\mathcal{G}$ & $\mathcal{R}$ \\
\midrule
canonical & Base & 50.16 & 53.65 & 75.30 \\
$20260730$ & Base & 49.69 & 52.24 & 73.75 \\
$20260731$ & Base & 49.70 & 53.67 & 75.40 \\
\midrule
mean$\pm$std & Base & \pmcell{49.85}{0.27} & \pmcell{53.19}{0.82} & \pmcell{74.81}{0.93} \\
\midrule
canonical & HC & 52.41 & 65.20 & 81.03 \\
$20260730$ & HC & 51.98 & 63.46 & 79.74 \\
$20260731$ & HC & 52.07 & 65.24 & 81.80 \\
\midrule
mean$\pm$std & HC & \pmcell{52.15}{0.22} & \pmcell{64.63}{1.02} & \pmcell{80.85}{1.04} \\
\bottomrule
\end{tabular}
\caption{SAME on \revised{FLEX} under the canonical task order and two random
permutations. Base is native soft routing; HC is the plug-in CIL router.
$\mathcal{G}$ and $\mathcal{R}$ are in percent. Std is the sample standard
deviation ($n{=}3$).}
\label{tab:flex_same_task_order}
\end{table}

Table~\ref{tab:flex_same_task_order} summarizes the three orders.
MacroScore varies little: Base $49.85\pm0.27$ (range $0.47$) and HC
$52.15\pm0.22$ (range $0.43$).
Over the same orders, $\mathcal{G}$ and $\mathcal{R}$ stay within about two
percentage points (Base ranges $1.43$ in $\mathcal{G}$ and $1.65$ in
$\mathcal{R}$; HC ranges $1.78$ and $2.06$).
Relative to Base, HC improves $\mathcal{M}$ by about $+2.3$ on every order,
with higher $\mathcal{G}$ and $\mathcal{R}$.
Under this protocol, task order barely affects performance on SAME Base and
SAME+HC, indicating robustness of the original soft-routing method and of the
embedded CIL router.
\FloatBarrier

\subsection{Modal Fusion}
\label{sec:supp_modal_fusion}

This experiment tests whether \revised{FLEX} effectively weakens textual
fingerprints, preventing a router from identifying tasks primarily through
instruction wording. In the original DISCO study, Sec.~5.3,
Table~3(a) compares identity-token extraction from text, images, and their
combination; text-only extraction performs best, while adding or replacing it
with visual features degrades performance~\citep{disco}. The authors attribute
this behavior to greater cross-task similarity in the image space than in the
instruction-text space. If \revised{FLEX} removes this textual shortcut, the
best DISCO configuration should consequently shift from text-only routing
toward genuinely multimodal routing.

Let $\lambda$ denote the image-routing weight and $1-\lambda$ the text-routing
weight. The image and text routers first produce soft task distributions
$\mathbf{p}^{v}$ and $\mathbf{p}^{q}$ using the native DISCO temperature
$T{=}0.05$, after which we fuse them at the probability level:
\begin{equation}
  \mathbf{p}
  =
  \lambda\mathbf{p}^{v}
  +
  (1-\lambda)\mathbf{p}^{q}.
\end{equation}
We evaluate $\lambda\in\{0,0.2,0.5,0.8,1.0\}$. All settings reuse the same
final DISCO checkpoint, 34 LoRA experts, and routing prototypes, while holding
the backbone, multimodal projector, decoding, and temperature fixed. The
text-only row reuses the formal DISCO Base result; the other four rows are
checkpoint-matched, inference-only evaluations.

\begin{table}[H]
\centering
\small
\begin{tabular}{lrrr}
\toprule
Image{:}Text & $\mathcal{M}$ & $\mathcal{G}$ & $\mathcal{R}$ \\
\midrule
$0{:}1$ (text only) & 50.32 & \textbf{43.90} & 63.95 \\
$0.2{:}0.8$ & 50.63 & 43.46 & 79.45 \\
$0.5{:}0.5$ & \textbf{50.79} & 42.80 & \textbf{83.86} \\
$0.8{:}0.2$ & 50.61 & 42.15 & 82.73 \\
$1{:}0$ (image only) & 49.90 & 41.71 & 74.65 \\
\bottomrule
\end{tabular}
\caption{DISCO modality-fusion sensitivity on the formal 34-task
\revised{FLEX}. All rows use soft, probability-level fusion with
$T{=}0.05$ and share the same trained experts and routing prototypes.}
\label{tab:flex_disco_modal_fusion}
\end{table}

Unlike on FCIT, the best DISCO setting on \revised{FLEX} is balanced
$0.5{:}0.5$ fusion rather than text-only routing. Relative to text only,
balanced fusion raises MacroScore from $50.32$ to $50.79$ and strict routing
match from $63.95\%$ to $83.86\%$. Text only retains the highest
ground-truth expert weight $\mathcal{G}$, showing that text remains useful,
but its much lower $\mathcal{R}$ indicates that text alone no longer
reliably separates the 34 tasks. Image-only routing also performs poorly,
with the lowest MacroScore of $49.90$, so neither modality is sufficient in
isolation.

The shift in the optimum from text-only routing on FCIT to balanced
image--text fusion on \revised{FLEX} provides direct evidence for the intended
benchmark design. Once task-specific textual fingerprints are weakened, the
router can no longer rely on formatting or wording shortcuts and must use
visual evidence to disambiguate tasks. \revised{FLEX} therefore turns routing
from textual-fingerprint matching into a genuinely multimodal task
identification problem.


\bibliography{aaai2027}